\documentclass{article}
\usepackage{arxiv}
\usepackage{graphicx}
\usepackage{bm}
\usepackage{amsmath}
\usepackage{newtxtext,newtxmath}
\usepackage{multirow}
\usepackage{url}

\title{A Cost-Efficient FPGA-Based CNN-Transformer using Neural ODE}

\author{
  Ikumi Okubo\\
  Keio University\\
  3-14-1 Hiyoshi, Kohoku-ku, Yokohama, Japan\\
  \texttt{okubo@arc.ics.keio.ac.jp} \\
  \And
  Keisuke Sugiura\\
  Keio University\\
  3-14-1 Hiyoshi, Kohoku-ku, Yokohama, Japan\\
  \texttt{sugiura@arc.ics.keio.ac.jp} \\
  \And
  Hiroki Matsutani\\
  Keio University\\
  3-14-1 Hiyoshi, Kohoku-ku, Yokohama, Japan\\
  \texttt{matutani@arc.ics.keio.ac.jp} \\
}

\begin{document}

\maketitle
\begin{abstract}
Transformer has been adopted to image recognition tasks and shown to outperform CNNs and RNNs while it suffers from high training cost and computational complexity.
To address these issues, a hybrid approach has become a recent research trend, which replaces a part of ResNet with an MHSA (Multi-Head Self-Attention).
In this paper, we propose a lightweight hybrid model which uses Neural ODE (Ordinary Differential Equation) as a backbone instead of ResNet so that we can increase the number of iterations of building blocks while reusing the same parameters, mitigating the increase in parameter size per iteration.
The proposed model is deployed on a modest-sized FPGA device for edge computing.
The model is further quantized by QAT (Quantization Aware Training) scheme to reduce FPGA resource utilization while suppressing the accuracy loss.
The quantized model achieves 79.68\% top-1 accuracy for STL10 dataset that contains 96$\times$96 pixel images.
The weights of the feature extraction network are stored on-chip to minimize the memory transfer overhead, allowing faster inference.
By eliminating the overhead of memory transfers, inference can be executed seamlessly, leading to accelerated inference.
The proposed FPGA implementation accelerates the backbone and MHSA parts by 34.01$\times$, and achieves an overall 9.85$\times$ speedup when taking into account the software pre- and post-processing.
The FPGA acceleration leads to 7.10$\times$ better energy efficiency compared to the ARM Cortex-A53 CPU.
The proposed lightweight Transformer model is demonstrated on Xilinx ZCU104 board for the image recognition of 96$\times$96 pixel images in this paper and can be applied to different image sizes by modifying the pre-processing layer.
\end{abstract}

\section{Introduction}\label{sec:intro}
Transformer-based \cite{Vaswani17} model with a sophisticated attention mechanism has been an active research topic.
ViT (Vision Transformer) \cite{Dosovitskiy21} is the first pure Transformer-based architecture for visual tasks and achieves promising results compared to existing CNN-based \cite{LeCun89} models.
However, training such models require large-scale datasets, e.g., JFT-300M \cite{Sun17}, limiting the application on resource-limited platforms.
Generally, this is due to the lack of inductive biases (e.g., locality and translational invariance) in the attention mechanism.
Thus, one of the solutions for alleviating the dependence on large datasets is to combine both convolution and attention mechanism \cite{Yuan21,Wu21,Trockman21,Aravind21,Namuk21,Dai21}.
Such hybrid models can achieve the state-of-the-art accuracy for small- or medium-scale datasets in a variety of tasks.

Among these models, BoTNet (Bottleneck Transformer Network) \cite{Aravind21} replaces spatial convolutions in the last three bottleneck blocks in ResNet \cite{He16} with the global self-attention (i.e., MHSA) and outperforms baseline models by a large margin.
We focus on BoTNet because it has shown an improved accuracy in object detection, instance segmentation, and image classification, demonstrating the benefits of including MHSA blocks in the CNN.

\begin{figure}[t]
    \centering
    \includegraphics[scale=0.5]{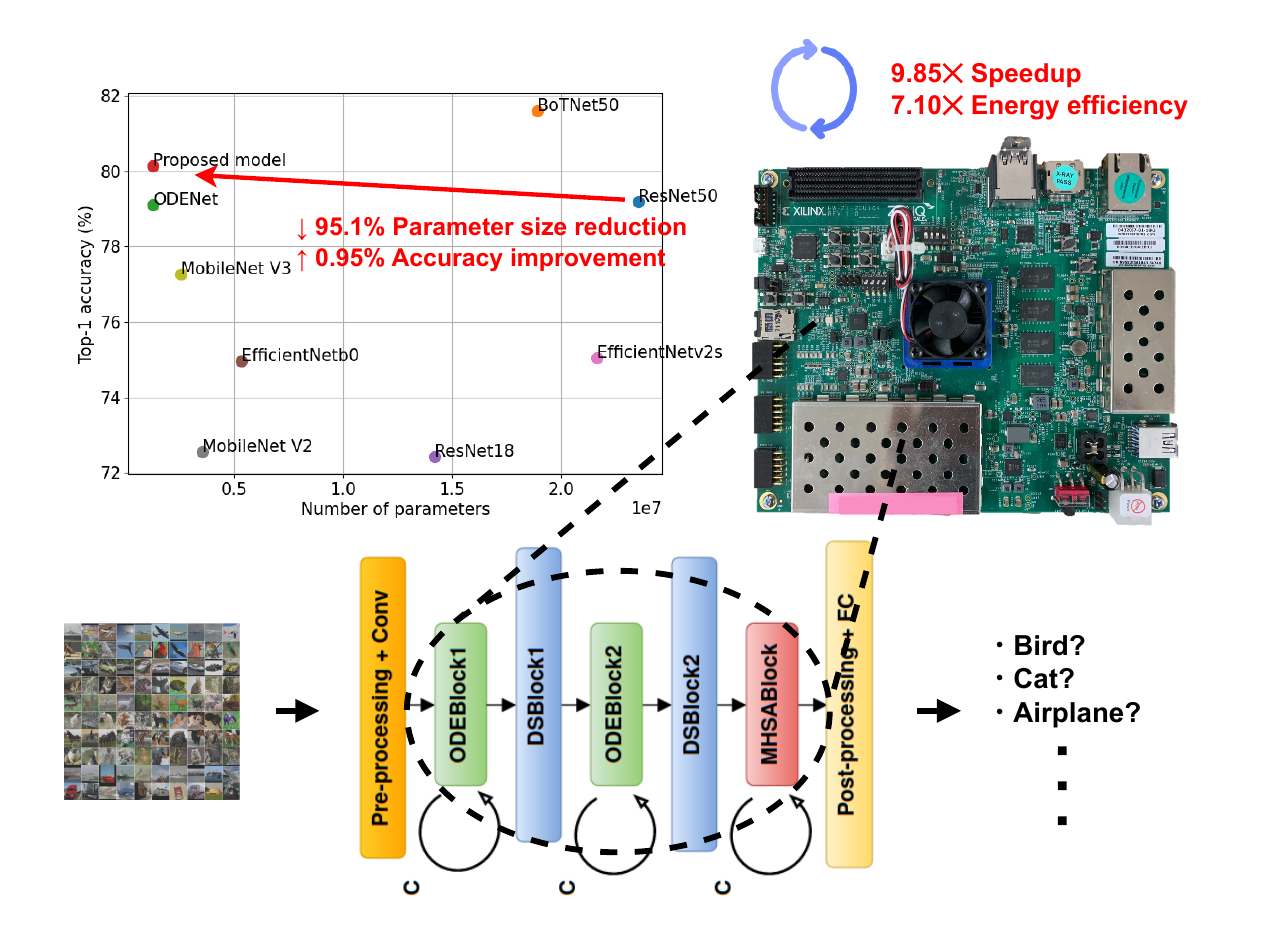}
    \caption{Overview of the proposed system}
    \label{fig:eye_catch}
\end{figure}

As mentioned above, the Transformer-based models are attractive in terms of accuracy while the computation cost is high (e.g., computation cost of MHSA grows with the square of input size $N$).
FPGA is becoming an attractive platform to accelerate DNNs due to its energy efficiency, reconfigurability, and short development period.
We thus aim to implement the proposed model including MHSA on a modest-sized FPGA such as Xilinx ZCU104.
In this paper, we propose a tiny Transformer-based model (see Fig. \ref{fig:eye_catch}) for resource-constrained FPGA devices by approximating the ResNet computation with Neural ODE \cite{Chen18}, which reformulates the forward propagation of ResNet blocks as a numerical solution of ODEs.
The number of parameters is greatly reduced because the forward pass through multiple ResNet building blocks is replaced by an iteration of the same single block, as shown in Fig. \ref{fig:eye_catch} (bottom).
We can thus implement a number of iterations of building
blocks while reusing the same parameters, significantly reducing the
parameter size compared to ResNets.

In our previous work \cite{Okubo23}, MHSA mechanism was implemented on the programmable logic part of the FPGA;
since a large part of the model was still performed on the host CPU, only a moderate performance improvement was achieved.
In contrast, this paper implements the entire feature extraction network of the proposed model on the FPGA to fully harness its computational capability.
Additionally, in this paper we choose to employ a lightweight quantization technique by considering the characteristics of Neural ODE, because the parameter reduction and model quantization play an important role to save the scarce on-chip memory and map the Transformer-based model onto such FPGAs.
Specifically, the model is quantized by QAT scheme instead of PTQ (Post Training Quantization) to further reduce FPGA resources while suppressing the accuracy loss.

The rest of this paper is organized as follows.
Sec. \ref{sec:related} presents related works and compares this work with them.
Sec. \ref{sec:prelim} provides the background on MHSA mechanism, Neural ODE, and quantization.
Sec. \ref{sec:design} presents the design of the proposed Transformer-based lightweight model by using Neural ODE.
Sec. \ref{sec:imple} implements the proposed model on the FPGA board using quantization technique to
accelerate the inference speed.
Sec. \ref{sec:eval} shows evaluation results of the proposed model in terms of parameter size and accuracy and those of 
the FPGA implementation in terms of resource utilization, execution time, and energy efficiency.
Sec. \ref{sec:conc} concludes this paper and discusses the future work.


\section{Related work} \label{sec:related}
\subsection{CNN and Transformer}\label{sec:rel_cnn_attention}

CNNs have been used for various image recognition tasks. For example, various CNN models are tested for detecting COVID-19 infection \cite{Das22,Kumar22}.
In addition, numerous Transformer-based models have been recently proposed for image recognition tasks.
ViT \cite{Dosovitskiy21} is a pure Transformer-based model for image recognition and achieves greater accuracy than CNN-based models with a large dataset (e.g., JFT-300M).
However, one major drawback of ViT is that it performs worse than CNN-based models 
when trained with small- to medium-scale datasets such as CIFAR-10/100 \cite{Krizhevsky09} and ImageNet \cite{Russakovsky15} due to the lack of intrinsic inductive biases.

CNNs have strong inductive biases such as spatial locality and translational invariance. 
The spatial locality is an assumption that each pixel only correlates to its spatially neighboring pixels.
While this assumption allows to efficiently capture local structures in image recognition tasks, it also comes with a potential issue of lower performance bound. 
On the other hand, the computation of the attention involves performing dot products between each pixel (i.e., all-to-all) in the feature map.
Therefore, the attention mechanism omits such inductive biases and aims to capture global structures \cite{Aravind21}.
In this case, 
the model has to be trained with large datasets to capture the important relationships between different parts of each input image, which may or may not be spatially close to each other.

Therefore, hybrid models that combine MHSA and convolution have been proposed \cite{Yuan21,Wu21,Trockman21,Aravind21,Namuk21,Dai21}.
BoTNet \cite{Aravind21} is introduced as a simple modification to ResNet, which replaces a part of convolutions with MHSAs, and has shown an improved accuracy in object detection, instance segmentation, and
image classification.
The experimental results in \cite{Aravind21} show accuracy improvements with larger image size and scale jitter, 
demonstrating the benefits of including MHSA blocks in the model.
By analyzing the relationship between convolution and MHSA,
it is shown that CNN tends to increase the dispersion of feature maps while MHSA tends to decrease it \cite{Namuk21}.
Considering these characteristics, AlterNet \cite{Namuk21} is proposed to suppress the dispersion of feature maps by adding MHSA to the final layer of each stage in ResNet.
It is shown that MHSA contributes to an improved accuracy and a flat and smooth loss surface, thereby increasing the model's robustness; as a result, AlterNet outperforms existing models on small datasets.

Recent studies have introduced hybrid models that combine CNNs and Transformers to address specific tasks in image processing, demonstrating superior accuracy.
For the task of single image deraining (i.e., removing degradations caused by rain from images), hybrid model that combines CNNs and Transformers is proposed in \cite{Chen23}.
A hybrid network combining CNNs and Transformers is proposed for lightweight image super-resolution in \cite{Fang22}.
For the purpose of denoising low-dose CT images, a hybrid model of CNNs and Transformers is introduced in \cite{Yuan23}.
These studies demonstrate the effectiveness of hybrid models that leverage the local feature extraction capability of CNNs and the long-range dependency modeling ability of Transformers.
These hybrid approaches offer a new direction by utilizing the strengths of different models to complement each other's weaknesses, thereby enhancing performance on specific tasks.

Many attention variants have been proposed to address the high computational cost of MHSA, which grows with the square of input size $N$.
In Reformer \cite{Kitaev20}, LSH (Locality Sensitive Hashing) technique is employed to efficiently retain regions with high attention scores, thereby reducing the computation complexity from $O(N^2)$ to $O(N \mathrm{log}N)$.
Other methods propose the Transformer with linear complexity $O(N)$. 
In linear Transformer \cite{Katharopoulos20}, the softmax operator is decomposed into two terms by employing a kernel method.
Linformer \cite{Wang20} is a low-rank factorization-based method.
Furthermore, in Flash Attention \cite{Dao22}, memory-bound Transformer processing is accelerated by introducing an IO-aware algorithm on GPUs.
These methods are proposed to overcome the computation cost associated with lengthy texts for NLP (Natural Language Processing) tasks.
A similar challenge arises in the context of high-resolution image recognition tasks.
Swim Transformer \cite{Liu21} employs fixed-pattern and global memory-based methods to reduce the computation cost from $O(N^2)$ to $O(N)$.
By employing such methods, 
it is possible to reduce the computational cost of the attention mechanism without sacrificing accuracy.

\subsection{Reduction of model computational complexity}\label{ssec:reduce_comp}
Considering the application of AI models on edge devices, running the inference of large-scale networks on such devices is intractable due to resource limitations. 
Generally, various techniques such as pruning, quantization, and low-rank factorization are employed for model compression. 
Aside from these, Neural ODE \cite{Chen18} can be applied to ResNet-family of networks with skip connections to reduce the number of parameters while avoiding the accuracy loss. 
It is emerging as an attractive approach to tackle the high computational complexity of large networks. Specifically, Neural ODE formulates the propagation of ResNet as ODE systems,
and the number of parameters is reduced by turning a set of building blocks into multiple iterations of a single block, i.e., reusing the same parameters.

Several extensions to Neural ODE have been proposed in the literature.
Neural SDE (Stochastic Differential Equation) \cite{Liu19} incorporates various common regularization mechanisms based on the introduction of random noise. 
This model shows greater robustness to input noise, whether they are intentionally adversarial or not, compared to Neural ODE. 
In \cite{Finlay20}, training cost is significantly reduced without sacrificing accuracy by applying two appropriate regularizations to Neural ODE. 
In \cite{Yan19}, by considering the regularization, TisODE (Time-invariant steady Neural ODE) is proposed to further enhance the robustness of vanilla Neural ODE.
In \cite{Dupont19}, inherent constraints in Neural ODE are identified, and more expressive model called ANODE (Augmented Neural ODE) is proposed.
These approaches are crucial in achieving robustness, approximation capabilities, and performance improvements in the context of Neural ODE. 
While this paper incorporates Transformer architecture into a model
with vanilla Neural ODE as a backbone, the same idea can be applied to
these Neural ODE extensions as well, though such exploration is beyond
the scope of this paper.

\subsection{FPGA implementation of Transformer architecture}\label{sec:rel_transformer_fpga}
One of advantages of the Transformer is its ability to capture a
global correlation among feature maps, thanks to the attention
mechanism which can enhance the accuracy.
This capability not only enhances the model accuracy but also aligns
well with FPGA-based acceleration since the attention mechanism relies
on matrix multiplications.
This operation is inherently parallelizable as well as that in CNNs.
Thus, the attention mechanism can efficiently utilize the parallel
processing capability of FPGAs while improving the model accuracy
compared to the traditional CNN-based implementations.

There is ongoing research on FPGA implementation of Transformer to improve the inference speed.
In \cite{Peng21}, an FPGA acceleration engine that employs a column balanced block-wise pruning is introduced for customizing the balanced block-wise matrix multiplications.
FTRANS \cite{Li20} uses a cyclic matrix-based weight representation for Transformer model for NLP.
VAQF \cite{Sun22} proposes an automated method 
to implement a quantized ViT on Xilinx ZCU102 FPGA board that meets a target accuracy.
In ME-ViT \cite{Marino24}, ViT is implemented on Xilinx Alveo U200.
It adopts a single-load policy where model parameters are loaded just once and operations are centralized within a single processing element.

\subsection{Research gap and objectives}\label{sec:gap_obj}

In the midst of the research trend on Transformer and CNN hybrid models \cite{Yuan21,Wu21,Trockman21,Aravind21,Namuk21,Dai21,Chen23,Fang22,Yuan23}, there are few proposals for small-scale hybrid models aimed at edge computing.
Although tiny Transformer models \cite{Wyatt21,Barbieri23} have been proposed for edge devices, they are mostly for signal processing and not for image recognition tasks.
The objective of this paper is thus to demonstrate a lightweight hybrid model of CNN and attention mechanism which can be implemented on resource-limited FPGA devices.
To fill in the research gap toward the objective, this paper introduces the Neural ODE method for the hybrid model and demonstrates the benefits of our approach on the modest-sized FPGA device for edge computing.

FPGA implementations of Transformers \cite{Peng21,Li20,Sun22,Marino24} mentioned in Sec. \ref{sec:rel_transformer_fpga} do not 
utilize CNNs (i.e., using pure Transformer models) and thus suffer from the drawbacks of attention mechanism (e.g., huge training cost). 
In addition, they consider the acceleration of large-scale Transformer models (e.g., ViT), which often require high-end FPGAs or memory access overheads to load from external memory to BRAM
due to the large number of parameters.
The model size is a crucial factor for deployment on resource-limited edge devices. Especially,
on-chip memory is a precious resource since it can offer low-latency access while its size is limited.
In this paper we thus present a tiny Transformer model with one million level parameters that fits within a
modest-sized FPGA (e.g., ZCU104) for image classification tasks.
The parameters are stored on-chip to eliminate the memory transfer overhead, allowing efficient inference.

\subsection{Implementation challenges}\label{sec:impl_challenge}

In \cite{Okubo23}, a lightweight Transformer model with Neural ODE is proposed along with its FPGA-based design.
Since only an MHSA block is accelerated and the other part including Neural ODE blocks is still executed on a host CPU in \cite{Okubo23}, the design shows only a limited performance gain.
Contrary to that, in this paper, our new approach is to implement the entire feature extraction network to further optimize the inference speed and fully benefit from the computing power of FPGAs.
Another implementation challenge is that, although the model in \cite{Okubo23} is quantized by simply converting inputs/outputs and weights from floating- to fixed-point representation, such simple type conversions throughout the entire model would result in a significant degradation of accuracy \cite{Okubo23}.
To address this issue, in this paper, QAT is instead employed, which allows more aggressive bit-width reduction (e.g., 8-bit quantization) to save the on-chip memory while mitigating the risk of significant accuracy loss.

\section{Preliminaries}\label{sec:prelim}
\subsection{MHSA}\label{sec:mhsa}

\subsubsection{Attention mechanism}\label{sec:for_set}
The attention mechanism aims to capture a relationship between 
a set of queries $\mathbf{Q} = \left[ \mathbf{q}_1, \ldots, \mathbf{q}_M \right]^\top \in \mathbb{R}^{M \times D}$ and a feature map 
$\mathbf{X} = \left[ \mathbf{x}_1, \ldots, \mathbf{x}_N \right]^\top \in \mathbb{R}^{N \times D}$ to identify the region of interest in $\mathbf{X}$ for each query $\mathbf{q}_i$, 
where $N$ and $D$ denote the number of elements and the number of channels for each element, respectively. 

Given a query vector $\mathbf{q} \in \mathbb{R}^D$, its corresponding attention weight $\mathbf{a} \in \mathbb{R}^N$ is calculated as follows:
\begin{equation}\label{eq:set_x}
    [a_1,\dots,a_N] = \mathrm{Softmax}\left(\frac{\mathbf{q}^{\top}\mathbf{X}^{\top}}{\sqrt{D}} \right).
\end{equation}
An attention $a_i$ represents the relevance between the query $\mathbf{q}$ and the feature element $\mathbf{x}_i$.
By stacking Eq. (\ref{eq:set_x}), attentions for $M$ queries can be calculated in the form of matrix operation as follows: 
\begin{equation}\label{eq:set_q}
    \mathbf{A} = \mathrm{Softmax}\left(\frac{\mathbf{QX^{\top}}}{\sqrt{D}} \right) \in \mathbb{R}^{M \times N}.
\end{equation}
Note that the softmax operator in Eq. (\ref{eq:set_q}) is applied row-wise to $\mathbf{Q}\mathbf{X}^{\top}.$

\subsubsection{Self-attention}\label{sec:SA}
Self-attention is a special case of the attention mechanism, where an input query $\mathbf{Q}$ is a feature map itself.
It can learn how to generate better feature maps, by calculating a correlation between every pair of feature elements.
First, query, key, and value matrices $\mathbf{Q}$, $\mathbf{K}$, and $\mathbf{V}$ are computed from $\mathbf{X}$ $\in \mathbb{R}^{N \times D}$ using three learnable weights 
$\mathbf{W}^q, \mathbf{W}^k,$ and $\mathbf{W}^v \in \mathbb{R}^{D \times D}$ as follows:
\begin{equation}
    \mathbf{XW}^q = \mathbf{Q}\in \mathbb{R}^{N\times D}
\end{equation}
\begin{equation}
    \mathbf{XW}^k = \mathbf{K}\in \mathbb{R}^{N\times D}
\end{equation}
\begin{equation}
    \mathbf{XW}^v = \mathbf{V}\in \mathbb{R}^{N\times D}.
\end{equation}
An attention map $\mathbf{A}$ $\in \mathbb{R}^{N \times N}$ is computed by multiplying $\mathbf{Q}$ with $\mathbf{K}^{\top}$ similar to Eq. (\ref{eq:set_q}), namely:
\begin{equation}\label{eq:A}
    \mathbf{A} = \mathrm{Softmax}\left(\frac{\mathbf{Q}\mathbf{K}^{\top} }{\sqrt{D}} \right) \in \mathbb{R}^{N\times N}.
\end{equation}
For each query $\mathbf{q}_i \in \mathbb{R}^{D}$, its output is obtained by calculating the sum of values $\left\{ \mathbf{v}_1, \ldots, \mathbf{v}_N \right\} $
weighted by their attention scores (i.e., coefficients of the $i$-th row vector of $\mathbf{A}$) $\mathbf{a}_i = \left[ a_{i, 1}, \ldots, a_{i, N} \right] \in \mathbb{R}^N.$  
That is, the output of self-attention is calculated by taking an inner product between $\mathbf{A}$ and $\mathbf{V}$ as follows:
\begin{equation}\label{eq:SA} 
    SA\mathbf{(X)} = \mathbf{AV} = \mathrm{Softmax}\left(\frac{\mathbf{Q}\mathbf{K}^{\top}}{\sqrt{D}} \right)\mathbf{V} \in \mathbb{R}^{N\times D}.
\end{equation}
Unlike convolution operations, which only use a subset of input features (i.e., a local region of input features) to compute each output element, 
in the self-attention, all input elements will contribute to the output for each query $\mathbf{q}_i$.
The self-attention thus aggregates the input feature globally and is not affected by inductive biases.

\subsubsection{Positional encoding}\label{sec:positional_encoding}
Since the self-attention (Eq. (\ref{eq:SA})) is a set operation and is invariant to the random permutation of inputs, it causes a loss of the positional information.
For example, in Eq. (\ref{eq:set_q}), if $\mathbf{x}_i$ and $\mathbf{x}_j$ are swapped, then their corresponding output elements, i.e., $\mathbf{a}_i$ and $\mathbf{a}_j$, are simply swapped.
The model with this property, called equivariance \cite{Dosovitskiy21}, is not appropriate for structural data such as images, since the model should be aware of the spatial relation of the pixels.
To address this information loss, a positional encoding is often employed with the attention mechanism.
A position-specific vector $\mathbf{p}_i \in \mathbb{R}^D$ is added to the 
$i$-th input $\mathbf{x}_i \in \mathbb{R}^D$ via addition $\mathbf{x}_i + \mathbf{p}_i$ or concatenation $\left[ \mathbf{x}_i, \mathbf{p}_i \right]$.
The positional encoding is modeled as a deterministic function with a set of fixed hyperparameters or as a learnable function. In addition, either absolute or relative position encoding can be used.
Typically, Transformer \cite{Vaswani17} adopts 
the sinusoidal function, which is categorized as a fixed absolute encoding and is written as follows:
\begin{eqnarray}
	\mathbf{p}_i = \left\{ \begin{array}{ll}
		\sin{(\frac{i}{10000^{2j/D}})}  &\qquad(i=2j)\\
		\cos{(\frac{i}{10000^{2j/D}})}, &\qquad(i=2j+1)
    \end{array} \right.
\end{eqnarray}
where $j=1,\dots,D/2$.
On the other hand, several studies \cite{Dosovitskiy21,Liu21} propose to learn the relative positional encoding via MLP.
In \cite{Aravind21}, the authors show that the relative position leads to better accuracy than the absolute one.
Thus, a learnable relative encoding is used in this study.

\subsubsection{MHSA} \label{ssec:mhsa}
MHSA is an extension of the self-attention mechanism \cite{Vaswani17}, 
which is comprised of multiple self-attention heads to jointly learn different relationships 
between features.
First, each weight matrix $\mathbf{W}^q, \mathbf{W}^k,$ and $\mathbf{W}^v \in \mathbb{R}^{N \times D}$ is partitioned column-wise into $k$ submatrices 
(i.e., $\mathbf{W}$ = $\mathrm{concat}[\mathbf{W}_1;\mathbf{W}_2;\dots;\mathbf{W}_k]$), 
each of which is fed to a separate self-attention head.
Let $SA_i$($\mathbf{X}$) be the $i$-th self-attention head, and its output is calculated as follows:
\begin{equation}
    SA_i\mathbf{(X)} = \mathrm{Softmax}\left(\frac{\mathbf{Q}_i\mathbf{K}_i^{\top}}{\sqrt{D}} \right)\mathbf{V}_i \in \mathbb{R}^{N\times D_h},
\end{equation}
where $\mathbf{Q}_i = \mathbf{X} \mathbf{W}_i^q \in \mathbb{R}^{N \times D_h}$, $\mathbf{K}_i = \mathbf{X} \mathbf{W}_i^k \in \mathbb{R}^{N \times D_h}$, 
and $\mathbf{V}_i = \mathbf{X} \mathbf{W}_i^v \in \mathbb{R}^{N \times D_h}$.
Following this, the output of MHSA is represented by Eq. (\ref{eq:MHSA}).
While the output dimension of each head is arbitrary, it is usually set to $D_h = D / k$ such that all heads and weight submatrices are equal-sized.
\begin{eqnarray}\label{eq:MHSA}
    \begin{split}
        \mathrm{MHSA}\mathbf{(X)} = [SA_{1}(\mathbf{X});...;SA_{k}(\mathbf{X})] \in \mathbb{R}^{N\times kD_h}
    \end{split}
\end{eqnarray}

\subsection{Neural Ordinary Differential Equation (Neural ODE)}\label{ssec:neuralode}
ResNet is a widely used backbone architecture especially for image classification tasks; it introduces residual connections 
between CNN-based building blocks (\textbf{ResBlock}s) to address the vanishing gradient problem and allow training deeper networks with tens of layers.
Neural ODE \cite{Chen18} is considered as a continuous generalization of ResNet, which interprets the skip connection as a discrete approximation 
of the ODE.
Let $f(\bm{z}_i, \theta_i)$ denote the $i$-th ResBlock with an input $\bm{z}_i$ and parameters $\theta_i$.
The forward propagation of $N$ ResBlocks is formally written as recurrent updates of the hidden state $\bm{z}$ as follows:
\begin{equation}
  \bm{z}_{i + 1} = \bm{z}_i + f(\bm{z}_i, \theta_i), \quad i = 0, \ldots, N - 1
  \label{eq:resnet-skip-connection}
\end{equation}
where the first additive term corresponds to the skip connection.

By treating the layer index $i$ as a time point and $\bm{z}_i$ as a time-dependent parameter evaluated at $t_i$, 
in the limit of infinitely many blocks (i.e., $N \to \infty$ and $\Delta t \to 0$), Eq. (\ref{eq:resnet-skip-connection}) turns into an ODE with respect to $t$ as follows:
\begin{equation}
  \frac{d\bm{z}}{dt} = f(\bm{z}(t), t, \theta).
  \label{eq:resnet-ode}
\end{equation}
The solution $\bm{z}(t_1)$ at some time point $t_1 \ge t_0$ is obtained by integrating Eq. (\ref{eq:resnet-ode}):
\begin{eqnarray}
  \bm{z}(t_1) &=& \bm{z}(t_0) + \int_{t_0}^{t_1} f(\bm{z}(t), t, \theta) dt \\
  &=& \mathrm{ODESolve}(\bm{z}(t_0), t_0, t_1, f),
  \label{eq:resnet-ode-int}
\end{eqnarray}
where $\bm{z}(t_0)$ is an initial state (i.e., input to the ResBlock) and $\mathrm{ODESolve}$ is an arbitrary ODE solver such as Euler and Runge-Kutta methods.
In the Euler method, the time $t \in \left[ t_0, t_1 \right]$ is again discretized by $t_j = t_0 + j h$ with a small step $h = (t_1 - t_0) / N$, and $\bm{z}(t)$ is iteratively solved as follows:
\begin{equation}
  \bm{z}(t_{j + 1}) = \bm{z}(t_j) + h f(\bm{z}(t_j), t_j, \theta).
  \label{eq:resnet-ode-euler}
\end{equation}

The observation of Eqs. (\ref{eq:resnet-skip-connection}) and (\ref{eq:resnet-ode-euler}) reveals that the forward propagation of a single ResBlock amounts to one iteration of the Euler method, i.e., 
by exploiting the Neural ODE formulation, $C$ different ResBlocks can be merged into one block (\textbf{ODEBlock}) that computes Eq. (\ref{eq:resnet-ode-euler}) $C$ times as shown in Fig. \ref{fig:ODENet} (left).
As apparent in Eq. (\ref{eq:resnet-skip-connection}), $C$ ResBlocks require $C$ individual sets of parameters $\left\{ \theta_1, \ldots, \theta_C \right\}$; 
compared to that, an ODEBlock reuses the same parameters $\theta$ during the $C$ iterations, leading to the $C$-fold reduction of parameter size.
Neural ODE approach allows to effectively build a compressed and lightweight alternative of ResNet-based deep models, by replacing $N$ ResBlocks with $N/C$ ODEBlocks.
Similar to ResNets, models composed by a stack of ODEBlocks are referred to as ODENets in this paper.

\subsection{Quantization of the proposed model}\label{ssec:quantization}
While there are a number of quantization methods, in this paper a learnable lookup table (LLT) based method \cite{Wang22} is employed, considering its simplicity and low computational overhead.
LLT-based quantization is outlined in the following.

In LLT, weight parameters and activations (i.e., layer inputs) are quantized, which are here denoted as $w$ and $a$.
Note that weights and activations have different value ranges: $w \in \mathbb{R}$ and $a \in \mathbb{R}^+$.
The latter is always positive because it is an output of the ReLU.
Considering this, 
we scale and clip $w$ and $a$ to the ranges $[-1, 1]$ and $[0, 1]$, respectively, using the $\mathrm{clip}(.)$ function as follows:
\begin{eqnarray}
    \hat{w} &=& \mathrm{clip}(w/s_w) \\
    \hat{a} &=& \mathrm{clip}(a/s_a),
\end{eqnarray}
where $s_w$ and $s_a$ are learnable scaling parameters.
Using $\hat{w}, \hat{a}, s_w,$ and $s_a$ obtained from the above equations, 
the quantized weight $\bar{w}$ and activation $\bar{a}$ are derived as follows: 
\begin{eqnarray}
    \bar{w} &=& s_w \cdot \mathbb{Q}_w(\hat{w}, Q_w) \label{eq:quant_w} \\
    \bar{a} &=& s_a \cdot \mathbb{Q}_a(\hat{a}, Q_a), \label{eq:quant_a} 
\end{eqnarray}
where $Q_w$ and $Q_a$ are the numbers of possible values for $w$ and $a$, respectively.
In the case of $n$-bit quantization, $Q_w = 2^{n-1}-1$ and $Q_a = 2^{n}-1$ are satisfied.
Furthermore, 
$\mathbb{Q}_w(.)$ and $\mathbb{Q}_a(.)$ are the mapping functions from full-precision values (i.e., $\hat{w}$ and $\hat{a}$) to $n$-bit integers, which are realized by learnable LUTs (Look-Up Tables).

In the following, the quantization process for activations is described. Note that weights can be quantized in the same way.
The full-precision activation $\hat{a} \in [0, 1]$ is converted to its nearest discrete value $i / 2^n$, which is represented by a total of $2^n - 1$ learnable step functions.
Here, when $\bar{a}$ is within the range $[i/2^n, (i + 1)/2^n]$, it is mapped to either $i/2^n$ or $(i + 1)/2^n$ by the $i$-th step function $\mathrm{step}_i(.)$ ($i = 0, 1, \ldots, 2^n - 1$) as follows:
\begin{eqnarray}
	\mathrm{step_i}(\hat{a}) = \left\{ \begin{array}{ll}
        \frac{i}{2^n} &\qquad(\hat{a} < T_i)\\
        \frac{i+1}{2^n}, &\qquad(\hat{a} \geq T_i)
    \end{array} \right.
\end{eqnarray}
where $T_i$ is a learnable threshold associated with each step function.
$T_i$ takes one of values in $[(i + 1)/2^n K, (i + 2)/2^n K, \ldots, (i + K)/2^n K]$, where $K$ is referred to as granularity.
This function can be represented as an equivalent LUT with $K$ elements, where the first $(T_i \cdot 2^n K - i)$ elements are $i/2^n$ and the rest are $(i + 1)/2^n$.
Such $2^n$ LUTs are concatenated into a single integrated LUT of size $2^nK$, which is denoted as ``I-LUT'' in this paper.
The quantization process with I-LUT corresponds to the function $\mathbb{Q}_a(.)$ in Eq. (\ref{eq:quant_a}).
Specifically, this quantization 
is achieved by a simple lookup operation as follows. 
For an input $\hat{a}$, its corresponding index $\mathrm{idx}$ is obtained by multiplying the table size $2^n K$ and then rounding to the nearest integer as in Eq. (\ref{eq:idx}).
\begin{eqnarray}
    \mathrm{idx} &=& \mathrm{round}(\hat{a} \times 2^n K) \label{eq:idx}\\
    \bar{a} &=& s_a \cdot \mathrm{I}\text{-}\mathrm{LUT}[\mathrm{idx}] \label{eq:bara}
\end{eqnarray} 

\section{Design of the proposed model}\label{sec:design}

This section describes the design of the proposed model.
Specifically, we outline key design aspects of the proposed model as follows:
(1) the ResNet architecture is first reviewed and (2) ODENet is introduced as a memory-efficient alternative to ResNet.
Then, (3) the DSC (Depth-wise Separable Convolution) is applied to ODENet to further reduce parameters. 
Analogous to BoTNet, (4) MHSA is incorporated into ODENet to derive a hybrid architecture, 
and (5) the three modifications are applied to MHSA to make it more hardware-friendly while preserving the accuracy. 

(1)
ResNet consists of pre-processing, post-processing, and multiple stages for feature extraction as shown in Table \ref{tbl:architecture}.
Pre-processing converts input images into feature maps, and 
post-processing transforms feature maps into a vector with elements equal to the number of classes.
Each stage comprises several ResBlocks with the same structure.
Specifically, each ResBlock contains two sets of convolution (\textbf{Conv}) layer, batch normalization (\textbf{BatchNorm}) layer, and activation function (e.g., ReLU), which are cascaded sequentially.
In each stage, the first ResBlock is employed for downsampling, denoted as ``DS'', followed by a stack of standard ResBlocks.
The ResBlock for DS halves the width and height of a feature map while doubling the number of channels, 
i.e., it takes the feature map of size $(Ch, H, W)$ as input and produces an output of size $(2Ch, H/2, W/2)$ \footnote{It is expected that the DS block reduces the feature map size while extracting more complex features by increasing the number of channels as the layer goes deeper. This structure is commonly used in well-known CNNs such as VGG \cite{Simonyan15}, ResNet \cite{He16}, and U-Net \cite{Ronneberger15}. We thus follow this approach in our proposed model.}.
Note that the ResBlock for DS and standard ResBlock only differ in the configuration of the first convolution layer.
For example, a straightforward approach to achieve DS is to set the stride and output channels of the first convolution in the block to 2 and $2Ch$, respectively.

\begin{figure}[t]
    \centering
    \includegraphics[scale=0.5]{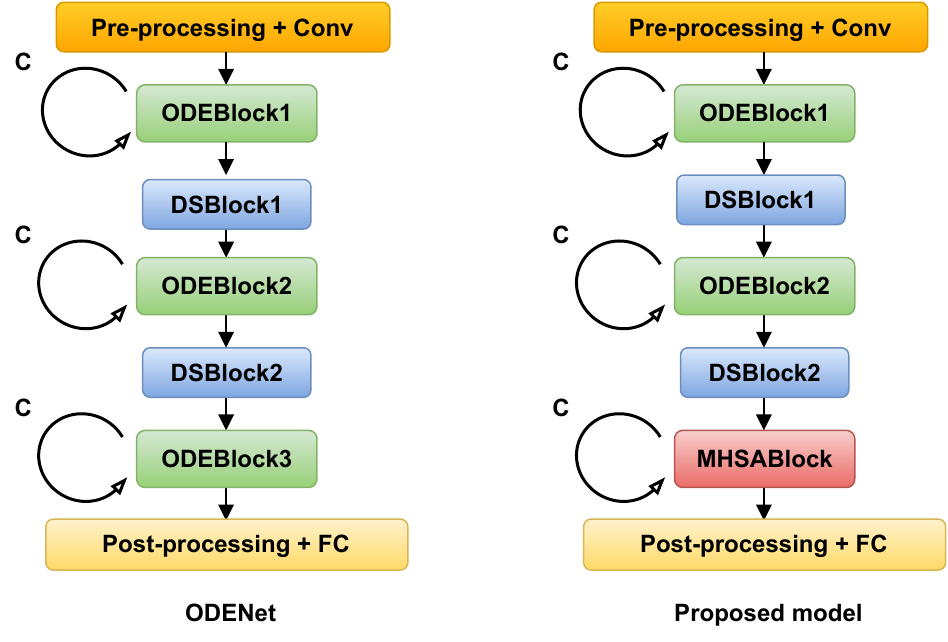}
    \caption{\textbf{Left:} ODENet, \textbf{Right:} the proposed model
    where only the third ODEBlock is replaced with MHSABlock}
    \label{fig:ODENet}
\end{figure}

(2)
As seen in Eq. (\ref{eq:resnet-skip-connection}), the size of feature map ($\bm{z}_0, \ldots, \bm{z}_{N-1}$) should remain the same during the forward propagation of $N$ ResBlocks, 
in order to merge them into a single ODEBlock. Since the first convolution in each stage changes the feature map size, a single ODEBlock cannot represent all ResBlocks in ResNet. 
Thus, we introduce special ODEBlocks corresponding to the ResBlocks for DS, referred to as \textbf{DSBlocks}.
Based on the above, a baseline architecture of ODENet \cite{Kawakami22} is shown in Fig. \ref{fig:ODENet} (left),
which consists of three ODEBlocks and two DSBlocks.
The computation of each ODEBlock is repeated $C$ times, while each DSBlock is executed only once during inference;
ODENet can be viewed as a deep network with 3$C$ + 2 blocks in total except that $C$ ODEBlocks in the same stage reuse the same parameters.
That is, ODENet in Fig. \ref{fig:ODENet} (left) can be regarded as an approximation of ResNet, which consists of three stages with each having $C$ ResBlocks.

\begin{figure}[t]
    \centering
    \includegraphics[scale=0.55]{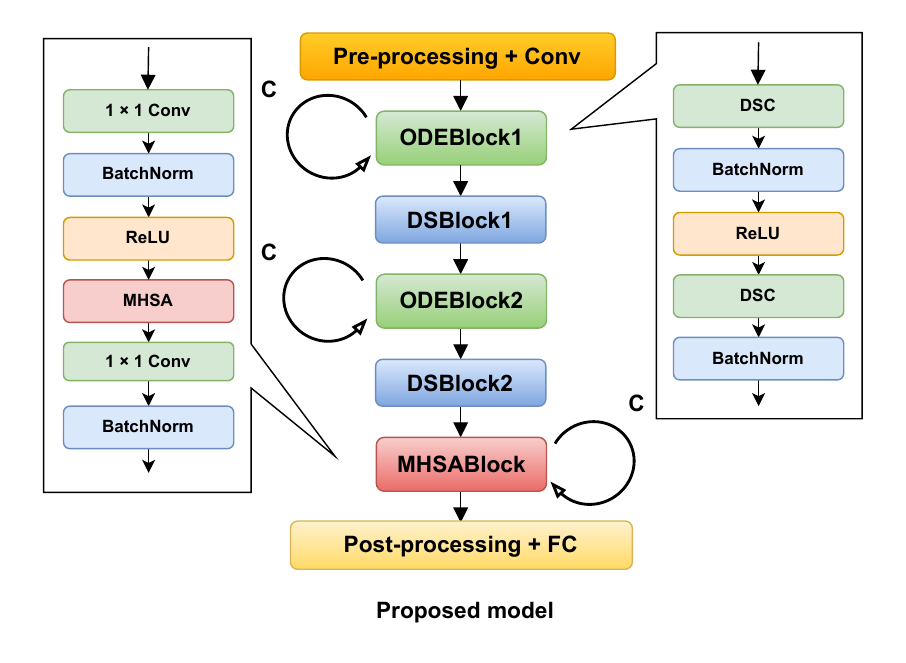}
    \caption{ODEBlock and MHSABlock in the proposed model}
    \label{fig:proposed}
\end{figure}

(3) 
Instead of the standard convolution, the lightweight CNN-based models \cite{mobilenets, xception} employ DSC, which performs convolution in the spatial and channel dimensions separately. 
While the standard convolution uses a total of $NMK^2$ parameters, DSC only uses $NK^2 + NM$ parameters, where $N, M,$ and $K$ denote the number of input channels, the number of output channels, and kernel size, respectively. 
Assuming $N, M \gg K$, DSC achieves approximately $K^2$ times parameter size reduction.
As shown in Fig. \ref{fig:proposed}, the proposed model also employs DSC in the ODEBlocks.

(4) The key idea of BoTNet is to replace 3$\times$3 convolutions in the last three ResBlocks with MHSA, referred to as \textbf{MHSABlocks}.
By applying such simple modifications, BoTNet has achieved a better accuracy than ResNet on ImageNet benchmark
while at the same time reducing the parameter size \cite{Aravind21}.
Since Neural ODE is a ResNet-like model, the idea of BoTNet can be applied to Neural ODE as well.
As depicted in Fig. \ref{fig:ODENet} (right), the final ODEBlock is replaced with MHSABlock, which has the same structure as in BoTNet.
While the model can adapt to arbitrary input image sizes by only modifying the pre-processing part (Fig. \ref{fig:ODENet}, top), in the FPGA implementation, the input image size of $96\times 96$ pixels is assumed.
As shown in Table \ref{tbl:eval_parameter},
it achieves 95.1$\%$ parameter size reduction compared to ResNet50, making it a tiny-sized Transformer model which is well-suited to deployment on resource-limited devices.

\begin{figure}[t]
    \centering
    \includegraphics[scale=0.65]{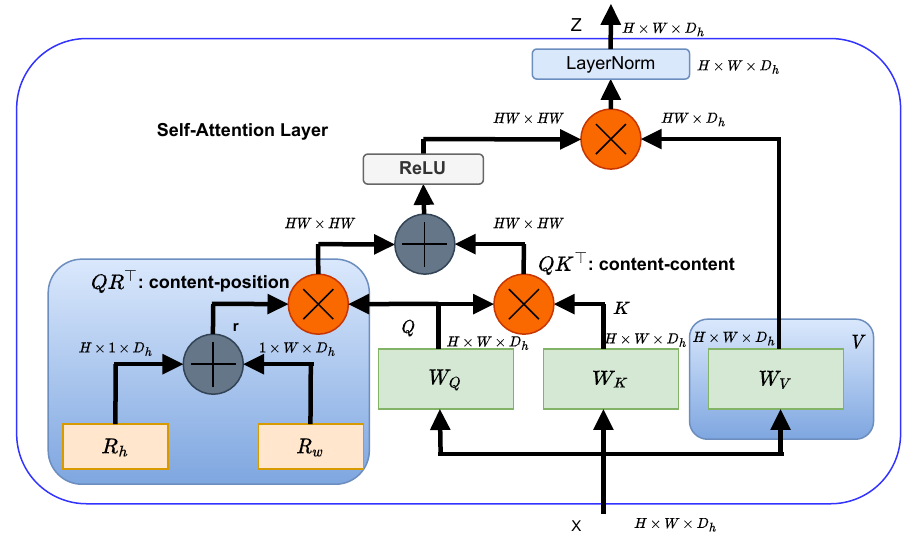}
    \caption{The modified MHSA implemented in FPGA}
    \label{fig:MHSA}
\end{figure}

(5) MHSABlock performs a slightly modified version of MHSA, which is shown in Fig. \ref{fig:MHSA}.
It uses a relative positional encoding instead of the absolute one as discussed in Sec. \ref{sec:mhsa}.
Following \cite{Bello19}, 
the relative positional encoding is applied to a query $\mathbf{Q}$ to obtain $\mathbf{Q} \mathbf{R}^\top$, where $\mathbf{R} = \mathbf{1}_w^\top \mathbf{R}_h + \mathbf{1}_h \mathbf{R}_w$.
$\mathbf{1}_w \in \mathbb{R}^{1 \times W \times 1}$ and $\mathbf{1}_h \in \mathbb{R}^{H \times 1 \times 1}$ are the vectors of all ones.
In this case, an attention $\mathbf{A}$ is computed as follows:
\begin{equation}\label{eq:A_rel}
    \mathbf{A} = \mathrm{Softmax}\left(\frac{\mathbf{Q}\mathbf{K}^{\top} + \mathbf{QR}^{\top}}{\sqrt{D_h}} \right), \mathbf{A}\in \mathbb{R}^{N\times N}.
\end{equation}

Another modification is to use ReLU\footnote{As an activation function, ReLU6 is also tested, and it shows a comparable accuracy to ReLU. 
In this paper, ReLU is used as well as \cite{Zhang21}.
} instead of softmax as an activation function.
As described in Sec. \ref{sec:mhsa}, relationships between query and key (i.e., logits)
are obtained by taking an inner-product of $\mathbf{Q}$ with $\mathbf{K}$, and then they are passed on to the softmax function
such that the attention weights $a_{i, 1}, \ldots, a_{i, N}$ for the $i$-th query sum to one.
An immediate advantage of this modification is that ReLU is hardware-friendly as it only consumes one comparator and one multiplexer.
According to \cite{Zhang21},
the accuracy of ReLU-based attention is comparable to that of the original softmax-based one. 
In addition, the attention weights become sparser,
which assists the analysis of information flow in the model.
Besides, LayerNorm \cite{Zhang19} is added at the end of MHSABlock to stabilize gradients and facilitate the convergence.

By combining the above two modifications, the MHSA is computed as follows:
\begin{equation}\label{eq:using_relu}
    \mathbf{A} = \mathrm{ReLU}\left(\frac{\mathbf{Q}\mathbf{K}^{\top} + \mathbf{QR}^{\top}}{\sqrt{D_h}} \right)
\end{equation}
\begin{equation}\label{eq:MHSA_inpl}
    \begin{split}
        \mathrm{MHSA(\mathbf{X})} 
        = \mathrm{LayerNorm}\left([SA_{1}(\mathbf{X});...;SA_{k}(\mathbf{X})]\right).
    \end{split}
\end{equation}

\section{Implementation}\label{sec:imple}
In this section, we describe the FPGA implementation of the proposed model.

We first discuss the FPGA implementation of the proposed model described in Sec. \ref{sec:design}.
The proposed system takes a software-hardware co-design approach.
The PL (Programmable Logic) part of the FPGA receives pre-processed feature maps from the PS (Processing System) part. 
Subsequently, the PL part processes these feature maps and returns the output to the PS part. Then, the PS part performs image classification by running the post-processing part.
Specifically, the feature extraction, i.e., the forward propagation from ODEBlock1 to MHSABlock (Fig. \ref{fig:ODENet}, right), is carried out on the PL part.
The pre- and post-processing parts are performed on the PS part with ARM Cortex-A53 CPU @1.2GHz.
The target FPGA board is the Zynq UltraScale+ MPSoC ZCU104.
The operating frequency of the PL part is set to 200MHz.

As described in Sec. \ref{sec:design}, the proposed model significantly reduces parameter size, 
and thus essential parameters for the feature extraction part can be 
stored in on-chip BRAM and URAM buffers. Only input and output data are transferred between the PL part and DRAM during inference.
In this case, because the data transfer amount is quite small, the performance of the proposed design is compute-bound. 
In the FPGA implementation, the model assumes the input image resolution of 96 $\times$ 96.
Note that the input image size has less impact on the model's parameter size because the feature map size does not affect the parameter size of convolution layers.

The FPGA implementation has two modes, which can be selected by control registers.
One mode involves the transfer of weights and LUTs for quantization from DRAM to the PL part.
This ensures that the necessary parameters are stored in on-chip buffers.
The other mode is for inference based on the trained parameters.

\subsection{Board-level implementation}
\begin{figure}[t]
    \centering
    \includegraphics[scale=0.47]{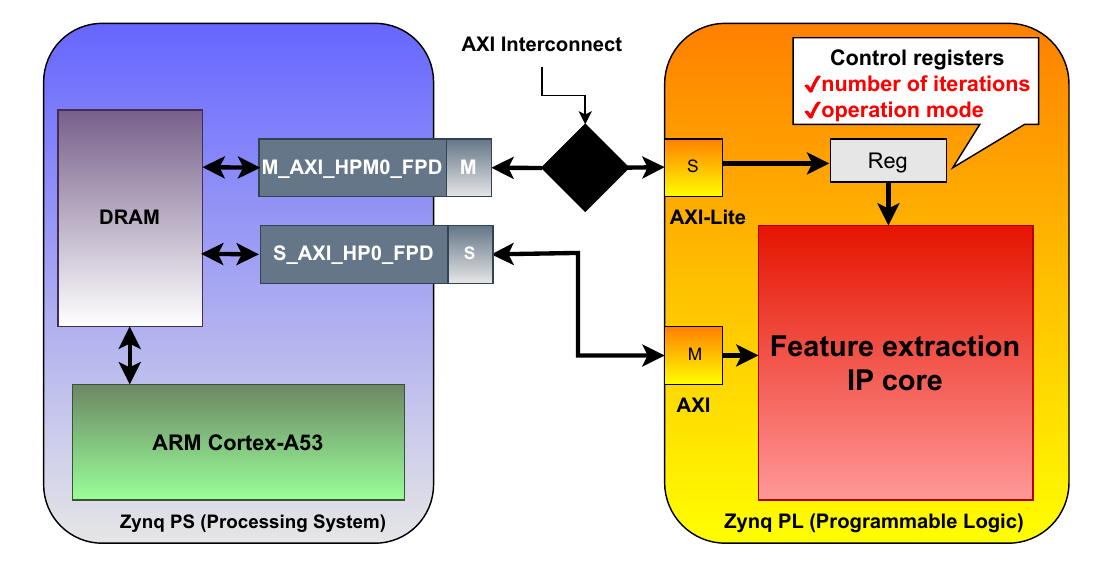}
    \caption{Board level implementation (Xilinx Zynq UltraScale+ MPSoC)}
    \label{fig:board_impl}
\end{figure}

Fig. \ref{fig:board_impl} depicts the block diagram of the proposed software-hardware co-design. The feature extraction part of the proposed model (i.e., from ODEBlock1 to MHSABlock) is designed as a custom IP core and implemented on the PL part.
The IP core transfers input/output feature maps as well as model parameters (e.g., LUTs for activations, quantized weights, and relative positional encoding) via the AXI master interface connected to the 128-bit High-Performance (HP0) port.
It also provides control registers for setting necessary configuration parameters such as the number of ODE iterations.
The PS part accesses these control registers through a 32-bit AXI-Lite interface  (Fig. \ref{fig:board_impl}, blue lines). The pre- and post-processing blocks are also executed on the PS part.

\begin{figure}[t]
    \centering
    \includegraphics[scale=0.65]{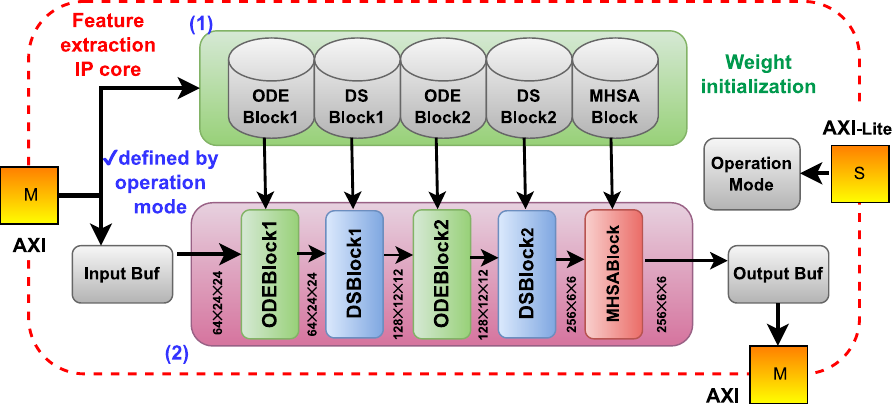}
    \caption{Overview of the proposed IP core}
    \label{fig:in_ipcore}
\end{figure}

\subsection{FPGA implementation of the proposed model}\label{ssec:imple_proposed_net}

The proposed model consists of three types of blocks: ODEBlock, DSBlock, and MHSABlock.
The proposed model is developed in C++ with Vitis HLS 2023.1 and implemented with Vivado 2023.1.
The proposed accelerator consists of a stream of blocks, as shown in Fig. \ref{fig:in_ipcore}.
Input samples are processed by separated blocks in order, namely ODEBlock1, DSBlock1, ODEBlock2, DSBlock2, and MHSABlock. 
These blocks use shared buffers for the communication between them.
Specifically, ODEBlock1 reads input features from the first buffer, computes output features, and stores them to the next buffer. 
It repeats these steps until all the input features are processed. 
Then, DSBlock1 reads input features from the second buffer and computes output features as in the previous block. 
These shared buffers are implemented with BRAMs and reused as much as possible to save the on-chip BRAM resources.
All layers in each block are parallelized along the output channel dimension by utilizing multiple DSP blocks.
We describe the implementation of ODEBlock, DSBlock, and MHSABlock in the following.

\subsubsection{Implementation of ODEBlock}\label{ssec:imple_odeblock}
\begin{figure}[t]
    \centering
    \includegraphics[scale=0.62]{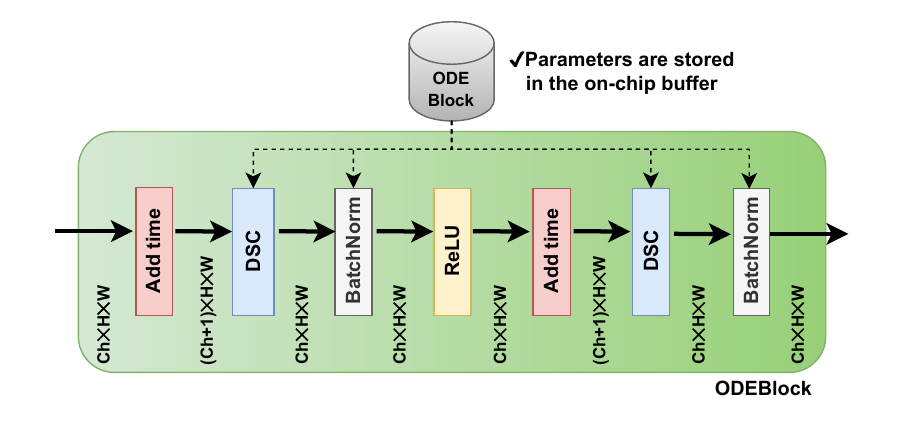}
    \caption{ODEBlock implementation}
    \label{fig:ODEBlock_impl}
\end{figure}

Fig. \ref{fig:ODEBlock_impl} illustrates the ODEBlock implementation. 
Since the residual connection of ResBlocks is replaced with ODEBlock, the shortcut connection is not included in the ODEBlock.
In the ODEBlock, the ordinary differential equation in Eq. (\ref{eq:resnet-ode-euler}) is solved numerically as described in Sec. \ref{ssec:neuralode}.
As shown in Table \ref{tbl:architecture}, ODEBlocks consist of several modules, such as Add time, DSC, BatchNorm, and ReLU.
The weight buffer size for each ODEBlock is determined by DSC and BatchNorm (Add time and ReLU do not have parameters).
DSC consists of point-wise and depth-wise convolution layers; thus, the weight buffer size for each DSC is $Ch_{in} \times K \times K + Ch_{out} \times Ch_{in}$, where
$Ch_{in}$ and $Ch_{out}$ represent the numbers of input and output channels of the feature maps, respectively.
BatchNorm requires two parameters precomputed by software for each channel; thus, the weight buffer size for each BatchNorm is $2 \times Ch$.
In our case (Table \ref{tbl:architecture}), the weight buffer sizes for ODEBlock1 and ODEBlock2 are 9,618 and 35,858, respectively.

As described in Sec. \ref{ssec:neuralode}, the number of iterations $C$ in ODEBlock is given as a hyperparameter and specified through a control register. 
The integration step size $h$ is determined as $1 / C$.
ODEBlock keeps track of a time variable $t$ ($t_j = t_0 + jh$ in Eq. (\ref{eq:resnet-ode-euler})) during iterations ($j = 0, \ldots, C - 1$). 
The module ``Add time'' in Table \ref{tbl:architecture} is to insert an additional channel filled with a time variable $t$ to the layer input; 
this increases the number of output channels by one.
In addition to on-chip buffers for weight parameters, extra buffers are allocated to retain the input to the block for skip connection and the output of each layer. 

\subsubsection{Implementation of DSBlock}\label{ssec:imple_dsblock}
\begin{figure}[t]
    \centering
    \includegraphics[scale=0.62]{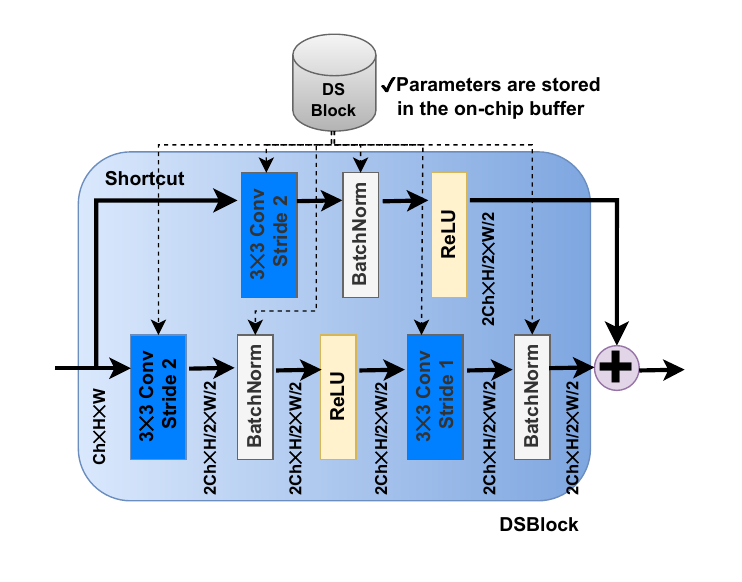}
    \caption{DSBlock implementation}
    \label{fig:DSBlock_impl}
\end{figure}

Fig. \ref{fig:DSBlock_impl} illustrates the DSBlock implementation.
The main purpose of DSBlock is to reduce the feature map size by half in the model.
DSBlocks are not iterated unlike ODEBlocks, and Neural ODE is not applied to the DSBlock in our implementation.
DSBlocks consist of several modules, such as convolution, BatchNorm, and ReLU.
The weight buffer size for each DSBlock is determined by the convolution and BatchNorm.
The weight buffer size for each convolution layer is $Ch_{out} \times Ch_{in} \times K \times K$.
The weight buffer size for each BatchNorm is $2 \times Ch$.
In our case (Table \ref{tbl:architecture}), the weight buffer sizes for DSBlock1 and DSBlock2 are 237,047 and 925,943, respectively.

\subsubsection{Implementation of MHSABlock}\label{ssec:imple_mhsablock}
\begin{figure}[t]
    \centering
    \includegraphics[scale=0.62]{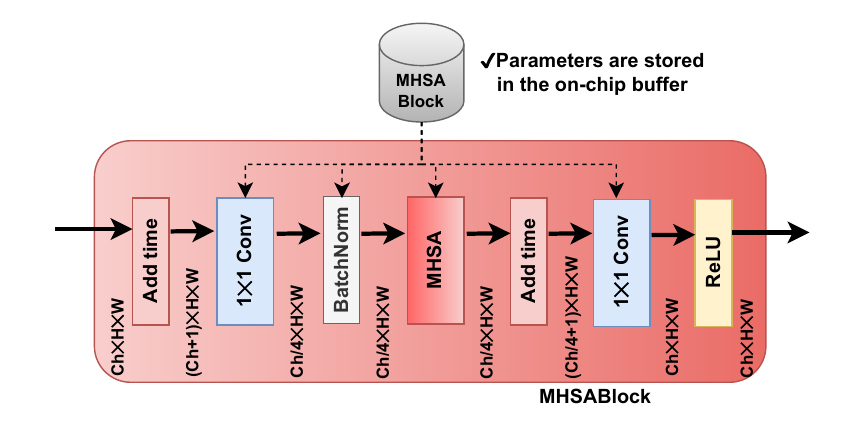}
    \caption{MHSABlock implementation}
    \label{fig:MHSABlock_impl}
\end{figure}

Fig. \ref{fig:MHSABlock_impl} illustrates the MHSABlock implementation. 
As in the ODEBlock, the shortcut connection is not included in the MHSABlock.
In the MHSABlock, a spatial convolution layer of the original ResNet building block is replaced with MHSA.
In addition to the convolution weights and batch normalization parameters, the three weights ($\mathbf{W}_k$, $\mathbf{W}_q$, $\mathbf{W}_v$) and pre-trained parameters for positional encoding ($\mathbf{R}_w$, $\mathbf{R}_h$) are stored on-chip.
MHSA (Fig. \ref{fig:MHSA}) is executed as follows:
(1) The query $\mathbf{Q}$, key $\mathbf{K}$, and product $\mathbf{QK}^{\top}$ are calculated from the input $\mathbf{X}$.
(2) Then, $\mathbf{QR}^{\top}$ is calculated and added to $\mathbf{QK}^{\top}$, where $\mathbf{R} = \mathbf{1}_w^\top \mathbf{R}_h + \mathbf{1}_h \mathbf{R}_w$ denotes the positional encoding.
(3) The result $\mathbf{QK}^{\top} + \mathbf{QR}^{\top}$ is divided by a scaling factor $\sqrt{D_h}$, and ReLU is applied to obtain an attention $\mathbf{A}$ (Eq. \ref{eq:using_relu}).
(4) The value $\mathbf{V}$ and product $\mathbf{A} \mathbf{V}$ (Eq. \ref{eq:SA}) are calculated from the input $\mathbf{X}$, and LayerNorm is applied to produce the final output of MHSA (Eq. \ref{eq:MHSA_inpl}).

The MHSABlock consists of multiple modules, such as Add time, point-wise convolution, BatchNorm, and MHSA.
MHSA has three weight parameters ($\mathbf{W}_q$, $\mathbf{W}_k$, $\mathbf{W}_v$) as well as those for the positional encoding and LayerNorm. 
The weight buffer size for $\mathbf{W}_q$, $\mathbf{W}_k$, and $\mathbf{W}_v$ is $3 \times Ch \times Ch$ in total and
that for the positional encoding is $Ch \times H \times W$.
LayerNorm also requires two parameters precomputed by software for each feature; the weight buffer size is $2 \times Ch \times H \times W$.
In our case (Table \ref{tbl:architecture}), the weight buffer size for MHSABlock is 64,433.

\begin{table}[t]
    \centering
    \caption{Architecture of the proposed model (for $96\times 96$ images)}
    \label{tbl:architecture}
    \begin{tabular}{c|c|c|l}
        \hline
        Block & Output & In/Out Ch & Layer \\ 
        \hline \hline 
        \multirow{2}{*}{Pre-processing} & \multirow{2}{*}{$24 \times 24$} & RGB $\rightarrow$ 64 & $7 \times 7$ Conv\\ \cline{4-3}
        & & 64 $\rightarrow$ 64 & $3 \times 3$ Max pool, stride 2 \\ \hline
        \multirow{7}{*}{ODEBlock1}&\multirow{7}{*}{$24 \times 24$} & 64 $\rightarrow$ 65 & Add time\\
        & & 65 $\rightarrow$ 64 &DSC\\
        & & 64 $\rightarrow$ 64 &BatchNorm\\
        & & 64 $\rightarrow$ 64 &ReLU\\
        & & 64 $\rightarrow$ 65 &Add time\\
        & & 65 $\rightarrow$ 64 &DSC\\
        & & 64 $\rightarrow$ 64 &BatchNorm\\\hline
        \multirow{8}{*}{DSBlock1} & \multirow{8}{*}{$12 \times 12$}& 64 $\rightarrow$ 128 &$3 \times 3$ Conv, stride 2\\
        & & 128 $\rightarrow$ 128 &BatchNorm\\
        & & 128 $\rightarrow$ 128 &ReLU\\
        & & 128 $\rightarrow$ 128 &$3 \times 3$ Conv, stride 1\\ 
        & & 128 $\rightarrow$ 128 &BatchNorm\\ \cline{3-4}
        & & 64  $\rightarrow$ 128 & Conv, stride 2 (short cut)\\
        & & 128 $\rightarrow$ 128 &BatchNorm\\
        & & 128 $\rightarrow$ 128 &ReLU\\ \hline
        \multirow{7}{*}{ODEBlock2}&  \multirow{7}{*}{$12 \times 12$} & 128 $\rightarrow$ 129 &  Add time\\
        & & 129 $\rightarrow$ 128  &DSC\\
        & & 128 $\rightarrow$ 128 &BatchNorm\\
        & & 128 $\rightarrow$ 128 &ReLU\\
        & & 128 $\rightarrow$ 129 &Add time\\
        & & 129 $\rightarrow$ 128 &DSC\\
        & & 128 $\rightarrow$ 128 &BatchNorm\\ \hline
        \multirow{8}{*}{DSBlock2} & \multirow{8}{*}{$6 \times 6$} & 128 $\rightarrow$ 256  & $3 \times 3$ Conv, stride 2\\
        & & 256 $\rightarrow$ 256 &BatchNorm\\
        & & 256 $\rightarrow$ 256 &ReLU\\
        & & 256 $\rightarrow$ 256 &$3 \times 3$ Conv, stride 1\\ 
        & & 256 $\rightarrow$ 256 &BatchNorm\\ \cline{3-4}
        & & 128 $\rightarrow$ 256 & Conv, stride 2 (short cut)\\
        & & 256 $\rightarrow$ 256 &BatchNorm\\
        & & 256 $\rightarrow$ 256 &ReLU\\ \hline
        \multirow{7}{*}{MHSABlock} & \multirow{7}{*}{$6 \times 6$} &256 $\rightarrow$ 257 &Add time \\
        & & 257 $\rightarrow$ 64 &$1 \times 1$ Conv\\
        & & 64 $\rightarrow$ 64 &BatchNorm\\
        & & 64 $\rightarrow$ 64 &MHSA\\
        & & 64 $\rightarrow$ 65 &Add time\\
        & & 65 $\rightarrow$ 256 &$1 \times 1$ Conv\\
        & & 256 $\rightarrow$ 256 &ReLU\\ \hline
        \multirow{2}{*}{Post-processing} & $1\times 1$ & \multirow{2}{*}{*} & Average pool\\ \cline{2-2}
        & * &  &Linear \\ \hline
    \end{tabular}
\end{table}

\subsection{Quantization of layers}\label{ssec:qol}
Furthermore, the design performs an LUT-based quantization as described in Sec. \ref{ssec:quantization}.
The accuracy with respect to different quantization levels is evaluated in Sec. \ref{ssec:quant_abration}.
Except the first and last layers in the proposed model, each fully-connected/convolution layer has two learnable LUTs for weights and activations.
The LUTs for weights are not necessary during inference, and thus only $n$-bit quantized weights are stored on-chip, which are computed by a host CPU at initialization.
On the other hand, the LUTs for activations are transferred to the PL part, since the activations are quantized during inference.
In addition, the scaling parameters $s_a$ and $s_w$ are transferred. 
The quantization of a single layer introduces additional $2^n K + 2$ parameters for one LUT and two scaling parameters, which increase the on-chip memory footprint by $(n \cdot 2^n K + 32 \cdot 2)$ bits.
On the other hand, the memory consumption for weights reduces from $32 N_p$ to $n N_p$ bits, where $N_p$ denotes the number of weight parameters in a layer.
$K$ is set to 9 as in \cite{Wang22}. In this case, the following inequality should be satisfied for the quantization to be effective.
\begin{equation}\label{eq:quant_bits}
    32 \times (2^n \times 9 + 2) < (32-n) \times N_p
\end{equation}
$N_p > 769$ holds when $n=8$ and $N_p > 385$ when $n=4$. In the proposed model, the quantization is 
applied to 
depth-wise/point-wise convolution layers present in ODEBlock1-2 and MHSABlock.
These convolution layers have 585 to 16,448 parameters and thus Eq. (\ref{eq:quant_bits}) holds in most cases except the smallest layer.
The memory reduction rate is calculated as follows:
\begin{equation}\label{eq:reduct_resource}
    \frac{32 \times N_p}{n \times N_p + n \times 2^n \times 9 + 32 \times 2}.
\end{equation}
For instance, the point-wise convolution layer in MHSABlock, which is the largest in the proposed model, achieves 3.51$\times$ and 6.99$\times$ reduction when $n = 4$ and $8$, respectively. 
In typical cases where $N_p \gg n$, Eq. (\ref{eq:reduct_resource}) can be approximated as $32 / n$.

The layer input is in full-precision and quantized according to Eqs. (\ref{eq:idx}) and (\ref{eq:bara}) to obtain $\bar{a}$. 
The layer computations such as convolutions are performed using the quantized $\hat{a}$ and $\bar{w}$.
While the weights are maintained as signed $n$-bit integers, it should be noted that during calculations, the $n$-bit weights are scaled by $s_w / 2^n$ to obtain the full-precision weight $\bar{w}$.

In our FPGA implementation, the quantization is selectively applied to some specific blocks by considering the accuracy and parameter size. 
In Sec. \ref{ssec:quant_abration}, ablation studies are conducted to investigate the effect of quantization and show that the model with DSBlocks and MHSABlock quantized (denoted as ``DS+Block3'' in Table \ref{tbl:final_acc}) strikes a good balance between the accuracy and parameter size.
Based on this result, 4-bit or 8-bit quantization is applied to DSBlocks and MHSABlock to implement dedicated IP cores.

\subsection{Local memory size}\label{ssec:local_mem}
The weight buffer size is 1,272,899 (9,618 + 35,858 + 237,047 + 925,943 + 64,433) in total as mentioned in Secs. \ref{ssec:imple_odeblock}--\ref{ssec:imple_mhsablock}.
It slightly differs from the number of parameters (1.26M) reported in Table \ref{tbl:eval_parameter} due to the additional parameters for the quantization (see Sec. \ref{ssec:qol}).
In addition, intermediate buffers are used to store output feature maps, and their size is 261,648 in total.
As a result, the total buffer size is 1,534,547.
In the DS+Block3 model, the quantized weight parameters (i.e., those in DSBlocks and MHSABlock) are represented as 4-bit or 8-bit integers.
The non-quantized parameters are represented as 16-bit fixed-point number format (integer part: 4-bit) in the PL part.
The LLT-quantized activations  are also represented as 4-bit or 8-bit integers.
The other activations are represented as 20-bit fixed-point number format (integer part: 10-bit) in the PL part.
As a result, the local memory sizes for the 4-bit and 8-bit quantized FPGA implementations are 1.23MB and 1.88MB, respectively.
Since 11Mb Block RAMs and 27Mb Ultra RAMs are available on the ZCU104 platform, these local memories can be placed on-chip.

\subsection{Degree of parallelism in each layer}
In our FPGA implementation, all the weight parameters used in the PL part are stored on-chip; 
thus, the performance bottleneck lies in the computation rather than the data transfer.
Thus, parallelization is applied to all the layers in Table \ref{tbl:architecture} to speed up the 
computation while fully utilizing the FPGA resources such as BRAM and DSP.
Based on simulation results from Vitis HLS, a higher degree of parallelism is applied to layers with longer execution time.
Table \ref{tbl:parallelism} shows the degree of parallelism and the number of cycles for each layer.
In this table, SC, DW, PW, and BNReLU stand for shortcut connection, depth-wise, point-wise, and BatchNorm and ReLU layers, respectively. 
Note that the same degree of parallelism is applied for both the 4-bit and 8-bit quantized implementations.

\begin{table}[t]
    \centering
    \caption{Degree of parallelism and execution cycles}
    \label{tbl:parallelism}
    \setlength{\tabcolsep}{3pt}
    \begin{tabular}{l|l|r|r}
        \hline
        Block & Layer & Parallelism & Latency (cycles)\\
        \hline
        \hline
        \multirow{3}{*}{ODEBlock1} & Add time + DWConv & 16 & 54,721\\
        & PW Conv & 64  & 54,720 \\ 
        & BNReLU & 2 & 18,439\\
        \hline
        \multirow{4}{*}{DSBlock1} & $3 \times 3$ Conv, stride 2 & 16 & 714,251\\
        & $3 \times 3$ Conv, stride 1 & 32 & 688,331 \\
        & $3 \times 3$ Conv, stride 2 (SC) & 8 & 781,483\\
        & BNReLU & 2 & 9,223 \\
        \hline
        \multirow{3}{*}{ODEBlock2} & Add time + DW Conv & 16 & 24,625\\
        & PW Conv & 64 & 37,160 \\ 
        & BNReLU & 2 &  9,223 \\
        \hline
        \multirow{4}{*}{DSBlock2} & $3 \times 3$ Conv, stride 2 & 16 & 688,907\\
        & $3 \times 3$ Conv, stride 1 & 32 & 676,235 \\
        & $3 \times 3$ Conv, stride 2 (SC) & 8 & 759,627 \\
        & BNReLU & 2 & 4,615 \\
        \hline
        \multirow{4}{*}{MHSABlock} & Add time + $1 \times 1$ Conv & 16 & 46,245 \\
        & BNReLU & 2 & 1,159 \\
        & MHSA & 16 & 269,448 \\
        & Add time + $1 \times 1$ Conv & 16 & 39,763 \\
        \hline
    \end{tabular}
\end{table}

\section{Evaluations}\label{sec:eval}
The proposed model is evaluated in comparison with ODENet \cite{Kawakami22}, BoTNet \cite{Aravind21}, ViT \cite{Dosovitskiy21}, 
and lightweight CNN-based architectures such as ResNet18\cite{He16}, EfficientNet \cite{Mingxing19}, EfficientNetV2 \cite{Mingxing21}, MobileNet V2 \cite{Sandler18}, and MobileNet V3 \cite{Qian21} in terms of parameter size and accuracy.
Furthermore, we conduct ablation studies to explore the trade-off between accuracy and resource utilization. 
We investigate the effectiveness of the newly added MHSABlock and the impact of LUT-based quantization on accuracy.
\subsection{Experimental setup}
\subsubsection{STL10 dataset}
The proposed model is a combination of CNN and attention mechanism. Since the attention mechanism aims to aggregate global information, it is expected to be beneficial in case of larger input images. 
As discussed in Sec. \ref{sec:rel_cnn_attention}, such a hybrid model can achieve a high accuracy even with small datasets compared to pure attention-based models (e.g., ViT). 
To demonstrate the benefit of the proposed hybrid model, its performance is compared with ViT-Base \cite{Dosovitskiy21} with a relatively small STL10 \cite{Coates11} dataset containing 10 object classes. STL10 contains 50k labeled images of $96\times 96$ pixels for training and 8k for testing. 
While CIFAR-10/100 \cite{Krizhevsky09} datasets are widely used for image classification models, they contain smaller images (e.g., $32\times 32$) than STL10.

\subsubsection{Model training}\label{sssec:model_training}
All network models considered here are implemented with Python 3.8.10 and PyTorch 1.12.1. 
The proposed model is built based on the code used in \cite{Kawakami22}. 
In the proposed model, DSC is applied to ODEBlocks while it is not applied to DSBlocks. In addition, by applying DSC to DSBlock1, DSBlock2, and both of them, 
we build three variant models, referred to as ``DS1DSC'', ``DS2DSC'',  and ``DS12DSC'', respectively.
In this paper, the same hyperparameters are used for training except the batch size; specifically, the batch size is set to 5 for the proposed model, and 128 for the others.
The number of epochs is set to 310, and SGD is used as an optimizer. 
The weight decay is set to $10^{-4}$ and the momentum to 0.9.
The cosine annealing with warm restart is used as a learning rate scheduler. 
The initial learning rate is set to 0.1, and the minimum learning rate is set to $10^{-4}$.
For data augmentation, the training images were randomly flipped, erased with a probability of 0.5, and jittered. 

\subsubsection{On-board execution}

The target FPGA board is the Zynq UltraScale+ MPSoC ZCU104.
As mentioned in Sec. \ref{sec:imple}, the PL part is in charge of ODEBlock1 to MHSABlock, while the PS part is in charge of the pre- and post-processing.
The PL part is reconfigured as the proposed IP core.
Pynq Linux 2.7 is running on the PS part.
Python 3.8.2 and PyTorch 1.13.1 are also running on the PS part.
The trained model and test dataset are downloaded to the PS part and then used by the successive blocks in the PL part.
The execution times of the PS and PL parts are measured by wall-clock time.

\begin{table}[t]
    \centering
    \caption{Parameter size and GFLOPs of the proposed and counterpart models}
    \label{tbl:eval_parameter}
    \begin{tabular}{l|r|r|c}
        \hline
        Model & Parameter size & GFLOPs &MHSA\\
        \hline \hline 
        ResNet50~\cite{He16} & 25.56M & 0.76 & \\
        BoTNet50~\cite{Aravind21} & \textbf{20.86M} & 0.74 &\checkmark \\
        \hline
        ODENet~\cite{Kawakami22} &  1.29M & 0.22 & \\
        Proposed model & \textbf{1.26M} & 0.21& \checkmark\\
        Proposed model (DS1DSC) & \textbf{1.08M}  & 0.19 & \checkmark\\
        Proposed model (DS2DSC) & \textbf{0.51M} & 0.19 & \checkmark\\ 
        Proposed model (DS12DSC) & \textbf{0.33M}  & 0.16 &\checkmark\\
        \hline
        ViT-Base~\cite{Dosovitskiy21} & 78.22M & 15.46 &\checkmark\\
        \hline
        ResNet18 & 11.69M & 0.34 & \\
        EfficientNetb0~\cite{Mingxing19} & 5.33M & 0.07 &\\
        EfficientNetv2s~\cite{Mingxing21} & 21.61M & 0.53 &\\
        MobileNet V2~\cite{Sandler18} & 3.54M & 0.06 &\\
        MobileNet V3~\cite{Qian21} & 2.56M & 0.01 &\\
        \hline
    \end{tabular}
\end{table}

\subsection{Evaluation results}
\subsubsection{Parameter size and computational cost}\label{ssec:Params}
Table \ref{tbl:eval_parameter} shows the parameter size and GFLOPs\footnote{GFLOPs highly depend on the input image size and those for 
STL10 (image size: 96$\times$96) reported in Table \ref{tbl:eval_parameter} are smaller than those for ImageNet (image size: 224$\times$224).}
 (giga floating-point operations) of the proposed and counterpart models with and without MHSA mechanism for the STL10 dataset. 
As shown, the proposed model uses the smallest number of parameters among these lightweight models, thanks to the benefits of Neural ODE.
BoTNet50 has 1.24$\times$ less parameters than ResNet50, because convolutions in the last three ResBlocks are replaced by MHSA. 
The proposed model achieves 14.96$\times$ reduction from BoTNet50 by combining MHSA and Neural ODE;
as described in Sec. \ref{sec:prelim}, the use of Neural ODE allows to compress $C$ ResBlocks into one ODEBlock, yielding $C$-fold reduction. In this case, $C$, i.e., the number of ODE solver iterations, is set to 10. 
By this model compression, the proposed model requires a high computational demand relative to parameter size.
From the table, it can be observed that GFLOPs per million parameters in ODENet and the proposed model (0.17--0.48) are mostly larger compared to the counterpart models (0.0039--0.20).
These models are thus well-suited for hardware accelerators in edge devices with limited memory capacity.

By employing DSC instead of the original convolution in DSBlocks, the parameter size can be further reduced by up to 3.86$\times$ (see DS1DSC, DS2DSC, and DS12DSC in Table \ref{tbl:eval_parameter}). 
Compared to the proposed model, ViT-Base is 61.96$\times$ larger, making it unsuitable for resource-limited devices due to its high computational cost. 
While EfficientNet and MobileNet are designed with the deployment on mobile devices in mind, they still have 2.02--17.12$\times$ more parameters than the proposed model, and also do not benefit from the attention mechanism.

\begin{table}[t]
    \centering
    \caption{Accuracy of the proposed and counterpart models}
    \label{tbl:eval_acc}
    \begin{tabular}{l|r|c}
        \hline
        Model & Top-1 accuracy (\%) & MHSA\\
        \hline \hline 
        ResNet50 & 79.20 & \\
        BoTNet & 81.60 ({\textbf{+2.40}})& \checkmark\\
        \hline
        ODENet & 79.11  &\\
        Proposed model & 80.15 ({\textbf{+1.04}})& \checkmark\\ 
        Proposed model (DS1DSC) &  80.05 &\checkmark\\
        Proposed model (DS2DSC) &  79.95 &\checkmark\\
        Proposed model (DS12DSC) &  79.29 & \checkmark\\
        \hline
        ViT-Base & 62.59& \checkmark\\
        \hline
        ResNet18 & 72.43&\\
        EfficientNetb0 & 74.96 &\\
        EfficientNetv2s & 75.05 &\\
        MobileNet V2 & 72.56 &\\
        MobileNet V3 & 77.26 &\\
        \hline
    \end{tabular}
\end{table}

\begin{figure}[t]
    \centering
    \includegraphics[scale=0.4]{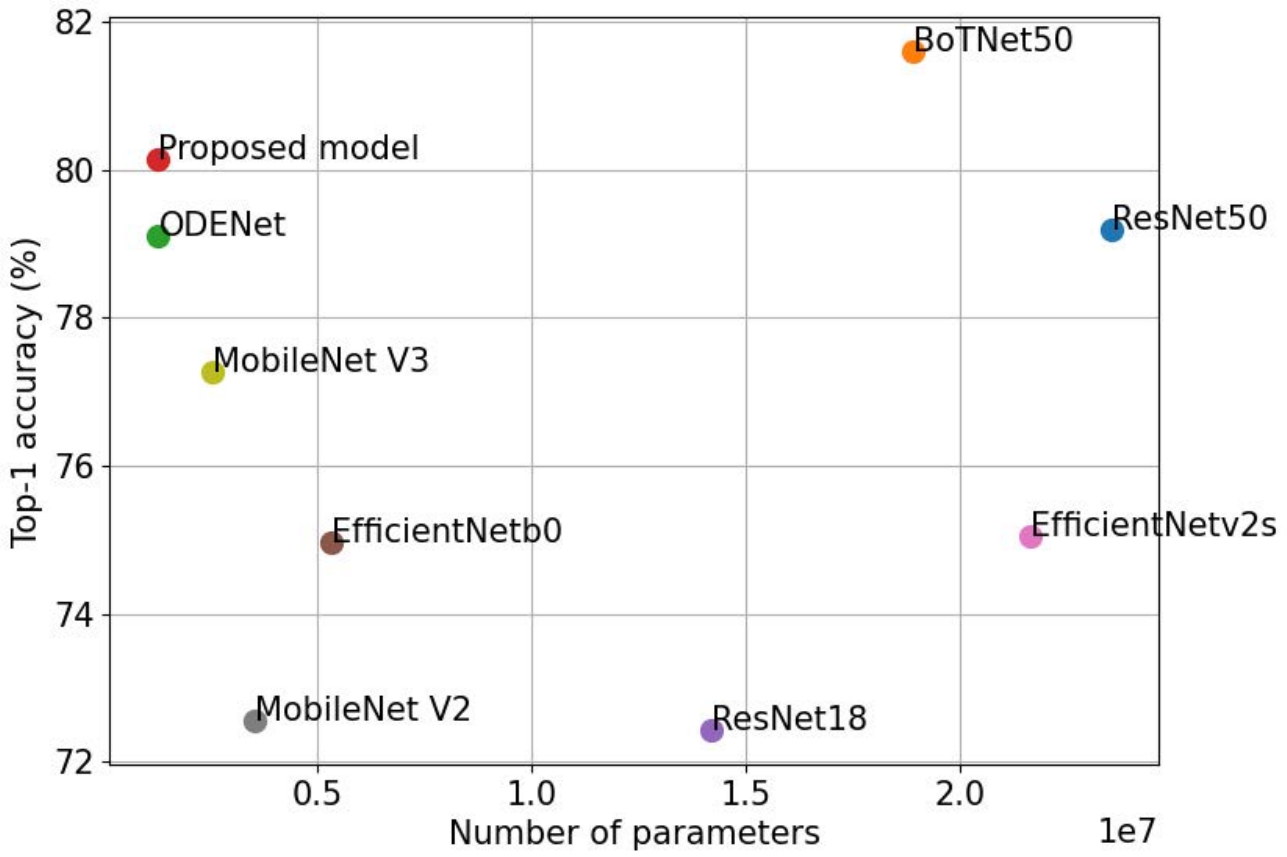}
    \caption{Number of parameters vs. top-1 accuracy}
    \label{fig:param_vs_acc}
\end{figure}

\subsubsection{Classification accuracy}
Table \ref{tbl:eval_acc} presents the top-1 accuracy of each model with STL10 dataset.
BoTNet shows 2.4\% higher accuracy than ResNet50, owing to the attention blocks placed at the end. Analogous to this, the proposed model achieves 1.04\% accuracy improvement over ODENet.
The proposed model which simplifies the architecture of BoTNet by introducing the concept of Neural ODE leads to a similar improvement on accuracy.
While the proposed model has 14.96$\times$ less parameters than BoTNet, it offers comparable accuracy and hence greatly improves the trade-off between accuracy and computational cost. 
In addition, it achieves 2.89--7.72\% better accuracy than EfficientNet and MobileNet series.
Fig. \ref{fig:param_vs_acc} visualizes the top-1 accuracy and parameter size of different models. The proposed model is in the top-left most corner, highlighting the better accuracy-computation trade-off.
Due to the scaling laws between the number of parameters and accuracy in AI models, models positioned in the upper left of the diagram tend to offer a favorable trade-off. 

\begin{table}[t]
    \centering
    \caption{Best accuracy of each combination of quantized blocks}
    \label{tbl:final_acc}
    \scalebox{1}[1]{
    \begin{tabular}{l|r|r|r} \hline
      Learning rate                    & Model   & 8-bit quant. (\%) & 4-bit quant. (\%) \\ \hline \hline
      [5e-3,1e-5]                      & Pre-trained & 80.15      & 80.15          \\ \hline
      \multirow{4}{*}{(a) [5e-3,1e-5]} & DS only & 80.97$\pm$0.08 & 81.48$\pm$0.21 \\
                                       & Block1  & 76.06$\pm$0.08 & 76.32$\pm$3.88 \\
                                       & Block2  & 78.18$\pm$0.17 & 77.49$\pm$0.12 \\
                                       & Block3  & 81.23$\pm$0.07 & 80.96$\pm$0.01 \\
                                       &DS+Block3& 81.23$\pm$0.27 & 78.84$\pm$0.11 \\ \hline
      \multirow{4}{*}{(b) [5e-4,1e-5]} & DS only & 80.83$\pm$0.15 & 81.06$\pm$0.05 \\
                                       & Block1  & 77.51$\pm$0.04 & 78.48$\pm$0.03 \\
                                       & Block2  & 78.71$\pm$0.03 & 78.69$\pm$0.00 \\
                                       & Block3  & 80.48$\pm$0.03 & 80.68$\pm$0.03 \\
                                       &DS+Block3& 80.48$\pm$0.17 & 80.03$\pm$0.13 \\ \hline
      \multirow{4}{*}{(c) [1e-4,1e-5]} & DS only & 79.70$\pm$0.05 & 80.27$\pm$0.20 \\
                                       & Block1  & 75.15$\pm$0.04 & 77.16$\pm$0.02 \\
                                       & Block2  & 77.25$\pm$0.00 & 77.95$\pm$0.01 \\
                                       & Block3  & 79.61$\pm$0.00 & 80.15$\pm$0.00 \\
                                       &DS+Block3& 76.53$\pm$0.20 & 79.10$\pm$0.18 \\ \hline
    \end{tabular}
    }
\end{table}

\subsubsection{Effects of quantization}\label{ssec:quant_abration}
To reduce the memory footprint and implement the proposed model on
modest-sized FPGAs, we want to apply quantization to as many layers as
possible.
On the other hand, since there is a trade-off between the degree of
quantization and accuracy, there is a risk of significant accuracy
degradation if quantization is applied to the model without careful
consideration.
Here, we conduct ablation studies to investigate the impact of
learning rates and quantized blocks.
Different combinations of the learning rates and quantized blocks are
evaluated.
In this experiment, 4-bit and 8-bit quantizations are applied to both
weights and activations for all the target blocks.
The initial and minimum learning rates for the cosine annealing
scheduler are set to [5e-3, 1e-5], [5e-4, 1e-5], and [1e-4, 1e-5].
DSC is applied to ODEBlocks only.

\begin{figure}[t]
    \centering
    \includegraphics[scale=0.3]{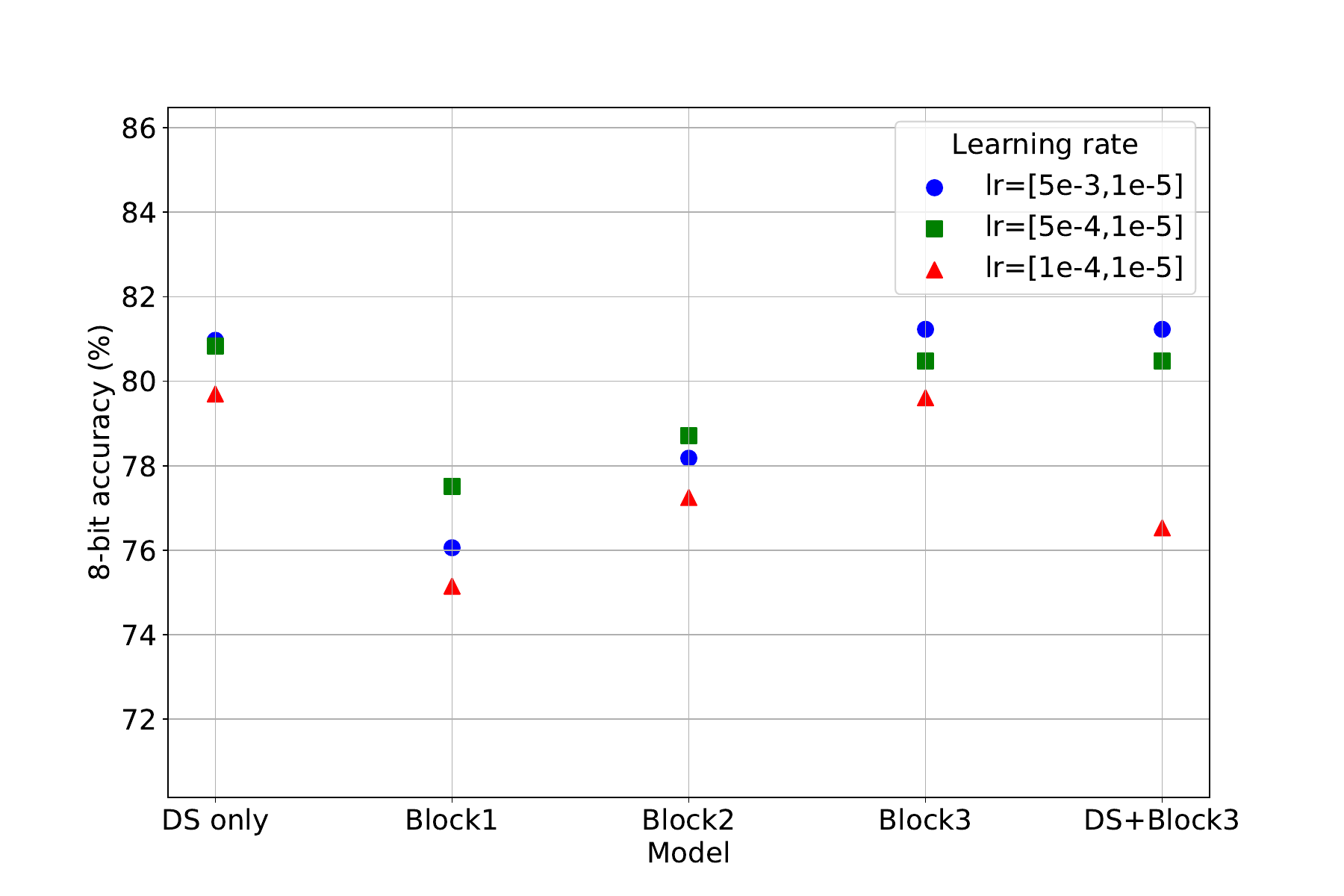}
    \caption{Best accuracy of each combination of quantized blocks (8-bit quant.)}
    \label{fig:8bit_acc}
\end{figure}

\begin{figure}[t]
    \centering
    \includegraphics[scale=0.3]{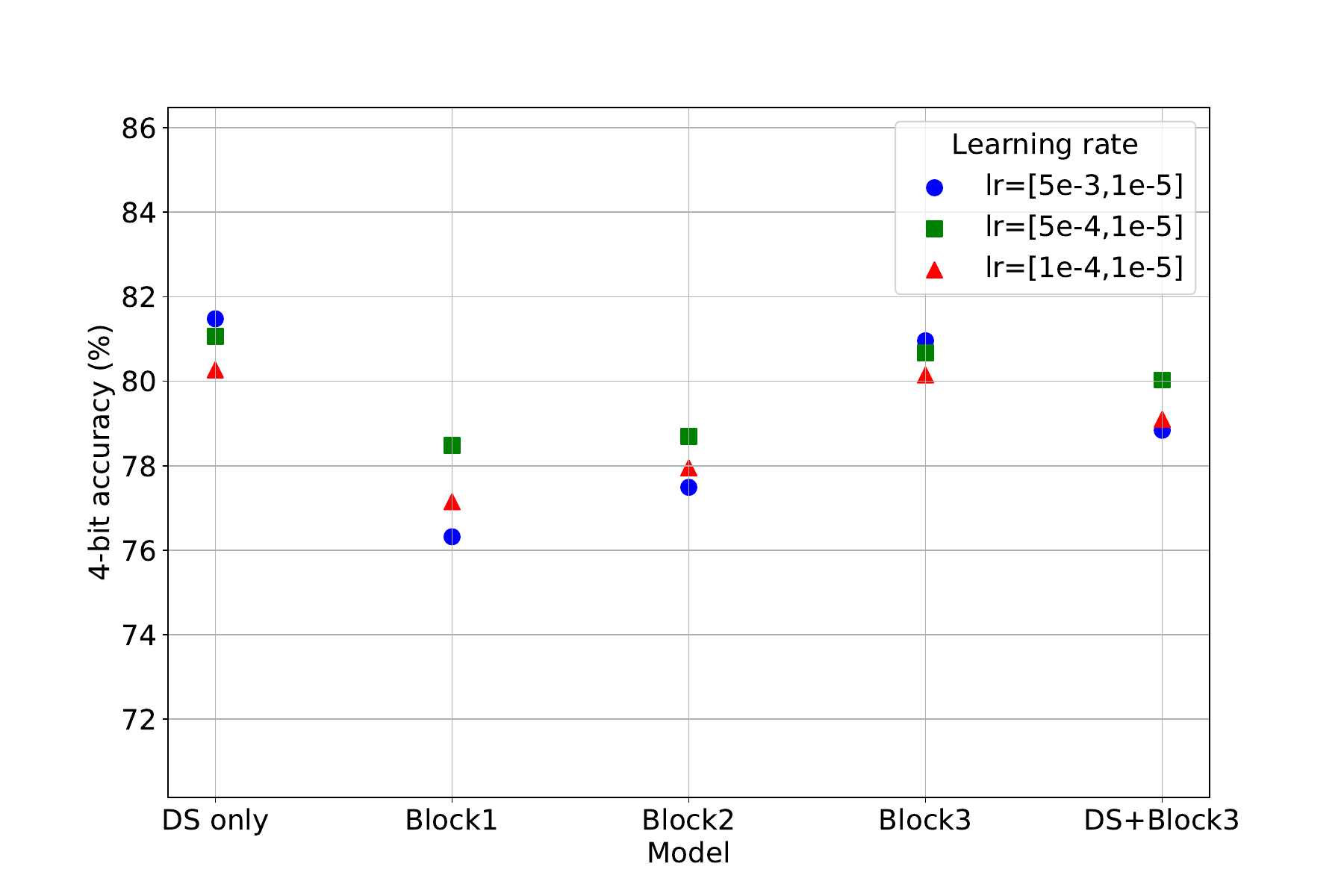}
    \caption{Best accuracy of each combination of quantized blocks (4-bit quant.)}
    \label{fig:4bit_acc}
\end{figure}

Table \ref{tbl:final_acc} shows the accuracy of the
proposed model when the 4-bit or 8-bit quantization is applied to
different blocks.
The accuracy without the quantization is also shown. 
In this table, ``DS only'' indicates that the quantization is applied
to DSBlocks only.
``Block1'', ``Block2'', and ``Block3'' refer to the proposed models
with the quantized ODEBlock1, ODEBlock2, and MHSABlock, respectively.
``DS+Block3'' indicates that the quantization is applied to DSBlocks
and MHSABlock.
Each model is trained with different random seeds and tested three
times.
Mean accuracies and their standard deviations are shown in the
table.
These results are also visualized in Figs. \ref{fig:8bit_acc} 
and \ref{fig:4bit_acc}.
For the 8-bit implementation, the best accuracy is achieved when the
quantization is applied to MHSABlock only and the learning rate
setting is [5e-3, 1e-5].
For the 4-bit implementation, the best accuracy is achieved when the
quantization is applied to DSBlocks only.
Considering that DS+Block3 gives almost the best accuracy for the
8-bit implementation while reducing the parameter size compared to DS
only and Block3, in this paper we implement the DS+Block3 model on the
FPGA, as evaluated below.

\subsubsection{FPGA resource utilization}
Table \ref{tbl:eval_resource_proposed} presents the FPGA resource
utilization of the DS+Block3 model for the 4-bit and 8-bit
implementations.
Thanks to the quantization and the 14.96$\times$ parameter size
reduction by Neural ODE, the feature extraction part of the proposed
model fits within the on-chip BRAM/URAM.
The 4-bit or 8-bit quantization is applied to both weights and
activations for DSBlocks and MHSABlock.
In the 4-bit and 8-bit implementations, BRAMs are preferentially
consumed for weights and activations, and URAMs are utilized for the
portions that exceed the BRAM capacity.
Thus, BRAM utilizations of the 4-bit and 8-bit implementations are
almost the same while their URAM utilizations differ.
As a result, compared to the 8-bit quantization, the URAM utilization
is further reduced by 17.7\% with the aggressive 4-bit quantization.
The number of DSP slices is the same in the 4-bit and 8-bit
implementations regardless of the quantized bit widths.
In this case, the benefit of the 4-bit implementation over the 8-bit one is the
parameter size reduction that can save the scarce on-chip memory
resources.
Although our 8-bit FPGA implementation already meets the resource
constraints of the Zynq UltraScale+ MPSoC ZCU104 platform, it would be
possible to further reduce the number of DSP slices for the 4-bit
implementation by adopting an aggressive DSP packing technique.

\begin{table}[t]
    \centering
    \caption{FPGA resource utilization of the proposed model}
    \label{tbl:eval_resource_proposed}
    \scalebox{1}[1]{
    \begin{tabular}{l|r|r|r|r|r} \hline
            Model & BRAM & DSP & FF & LUT & URAM\\ \hline \hline
            Available & 624 & 1,728 & 460,800 & 230,400 & 96\\ \hline
            \multirow{2}{*}{8-bit quant.} & 595& 206& 23,681 & 66,292 & 46 \\
                                                & 95.4$\%$ & 11.92$\%$ & 5.14$\%$ & 28.77$\%$ & 47.92$\%$ \\ \hline
            \multirow{2}{*}{4-bit quant.} & 589 & 206 & 24,143 &  66,843 & 29 \\
                                               & 94.4$\%$ & 11.92$\%$ & 5.24$\%$ & 29.01$\%$ & 30.21$\%$ \\ 
    \hline
    \end{tabular}
    }
\end{table}

\begin{table}[t]
    \centering
    \caption{Top-1 accuracy in software and FPGA board}
    \label{tbl:acc_in_fpga}
    \begin{tabular}{c|c|r|r} \hline
        Platform   & Model     & 8-bit quant. (\%) & 4-bit quant. (\%) \\\hline
        Simulation & DS+Block3 & 81.23$\pm$0.27    & 80.03$\pm$0.13 \\
        FPGA board & DS+Block3 & 79.68$\pm$0.37    & 76.97$\pm$0.36 \\
        \hline
    \end{tabular}
\end{table}

\begin{table}[t]
    \centering
    \caption{Execution time (ms) and GOPS of CPU and FPGA implementations}
    \label{tbl:eval_time}
    \begin{tabular}{l|c|r|r|r|r} \hline 
        Model &  & Mean & Max &  Std & GOPS \\ \hline \hline
        CPU (8-bit quant.) & PS & 1,596 & 1,676 &  12.73 & 0.13 \\ \hline
        CPU (4-bit quant.) & PS & 1,590 & 1,640 & 9.27 &  0.13 \\ \hline
         & PS+PL & 162.1 & 167.5 & 14.40 &  1.30 \\
         FPGA (8-bit quant.) & PS & 118.7 & 124.5 & 0.76 & 0.19 \\
         & PL & 43.44 & 43.45 & 0.02 & 4.33 \\ \hline
         & PS+PL & 161.8 & 170.3 & 18.60  & 1.30 \\
         FPGA (4-bit quant.)& PS & 118.6 & 127.6 & 1.06 & 0.19\\
        & PL & 43.33 & 43.34 & 0.03 & 4.34\\
    \hline
    \end{tabular}
\end{table}

\subsubsection{FPGA inference accuracy}
Here, our 4-bit and 8-bit FPGA implementations are executed on the ZCU104 platform and evaluated in terms of inference accuracy.
As mentioned in Sec. \ref{ssec:qol}, the DS+Block3 model is
implemented on the ZCU104 platform.
The 8-bit and 4-bit quantized models trained with learning rates of
[5e-3, 1e-5] and [5e-4, 1e-5], respectively, are used for the
implementation.
Note that only DSBlocks and MHSABlock are quantized in the DS+Block3 model. 
The parameters for the other remaining blocks are represented as float32 in the software, and fixed-point in the FPGA implementation.
To see this difference, the simulation and on-board results of the 4-bit and 8-bit implementations are compared in terms of inference accuracy.
The mean accuracies in Table \ref{tbl:acc_in_fpga}
demonstrate that the accuracy loss of the FPGA implementation is small
in the 8-bit case.
Note that if we can selectively implement the best parameters (among
three training trials) in Table \ref{tbl:final_acc} on the FPGA, the
maximum on-board accuracy in the 8-bit case is 80.20\%; in this case,
the on-board accuracy is comparable to that of the baseline
floating-point model (i.e., 80.15\%) which is denoted as
``Pre-trained'' in Table \ref{tbl:final_acc}.

\subsubsection{Execution time}

\begin{figure}[t]
    \centering
    \includegraphics[scale=0.35]{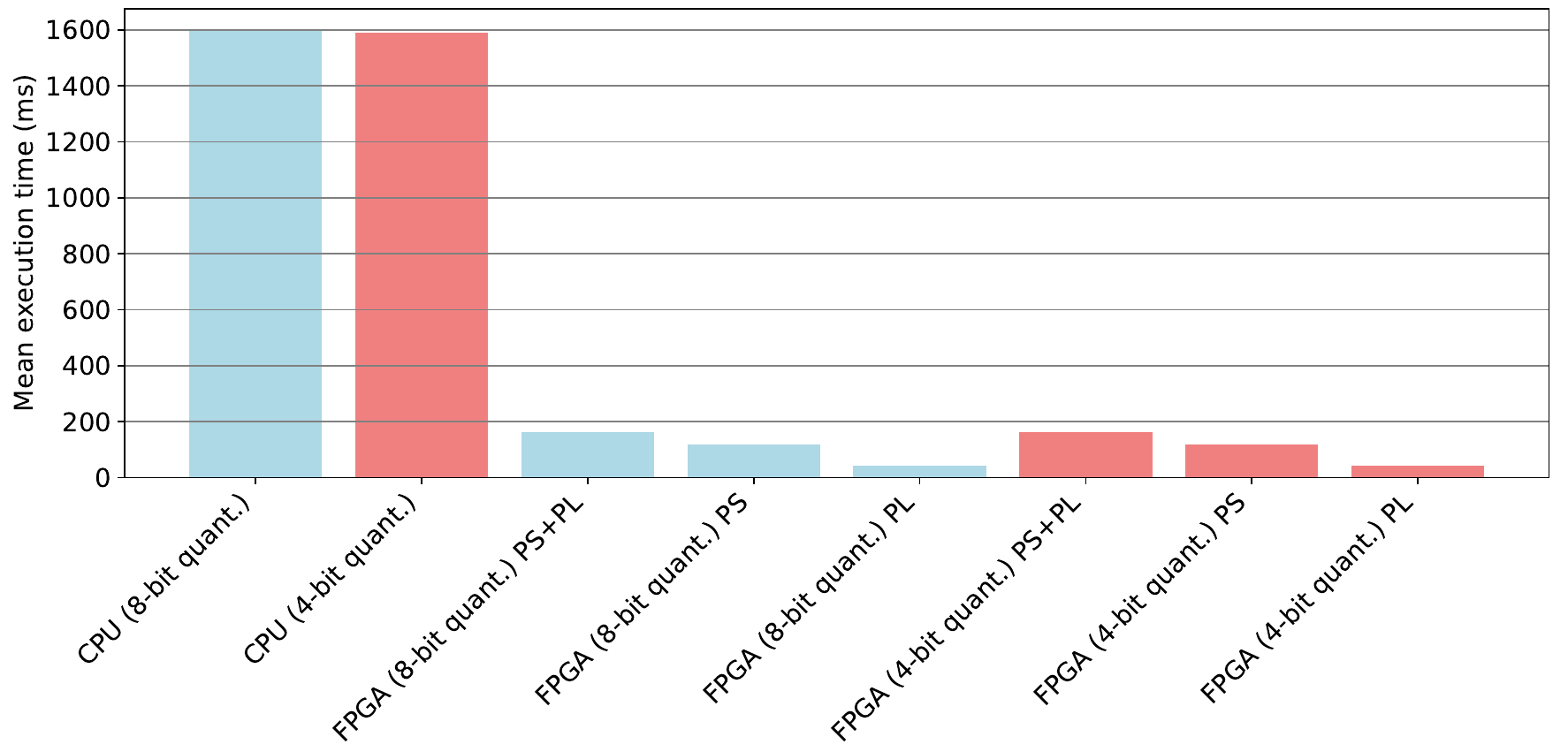}
    \caption{Execution time (ms) of CPU and FPGA implementations}
    \label{fig:exec_time}
\end{figure}

In this section, the performance of the FPGA implementation is compared with its software counterpart running on ARM Cortex-A53 CPU.
The software counterpart is implemented with PyTorch.
The 8-bit and 4-bit quantizations are applied to both implementations.
Table \ref{tbl:eval_time} presents a distribution of the execution time obtained by 100 runs.
The results are visualized in Fig. \ref{fig:exec_time}.
Please note that, in the FPGA implementation, only the pre- and post-processing part is executed on the PS part while the feature extraction part is handled in the PL part.
The total execution times (PS + PL) of the 8-bit and 4-bit FPGA implementations are 162.1ms and 161.8ms; 
thus, they achieve 9.85$\times$ and 9.83$\times$ speedups, respectively.
If we only consider the feature extraction part, they achieve 34.01$\times$ and 33.96$\times$ speedups, respectively.

\begin{figure}[t]
    \centering
    \includegraphics[scale=0.35]{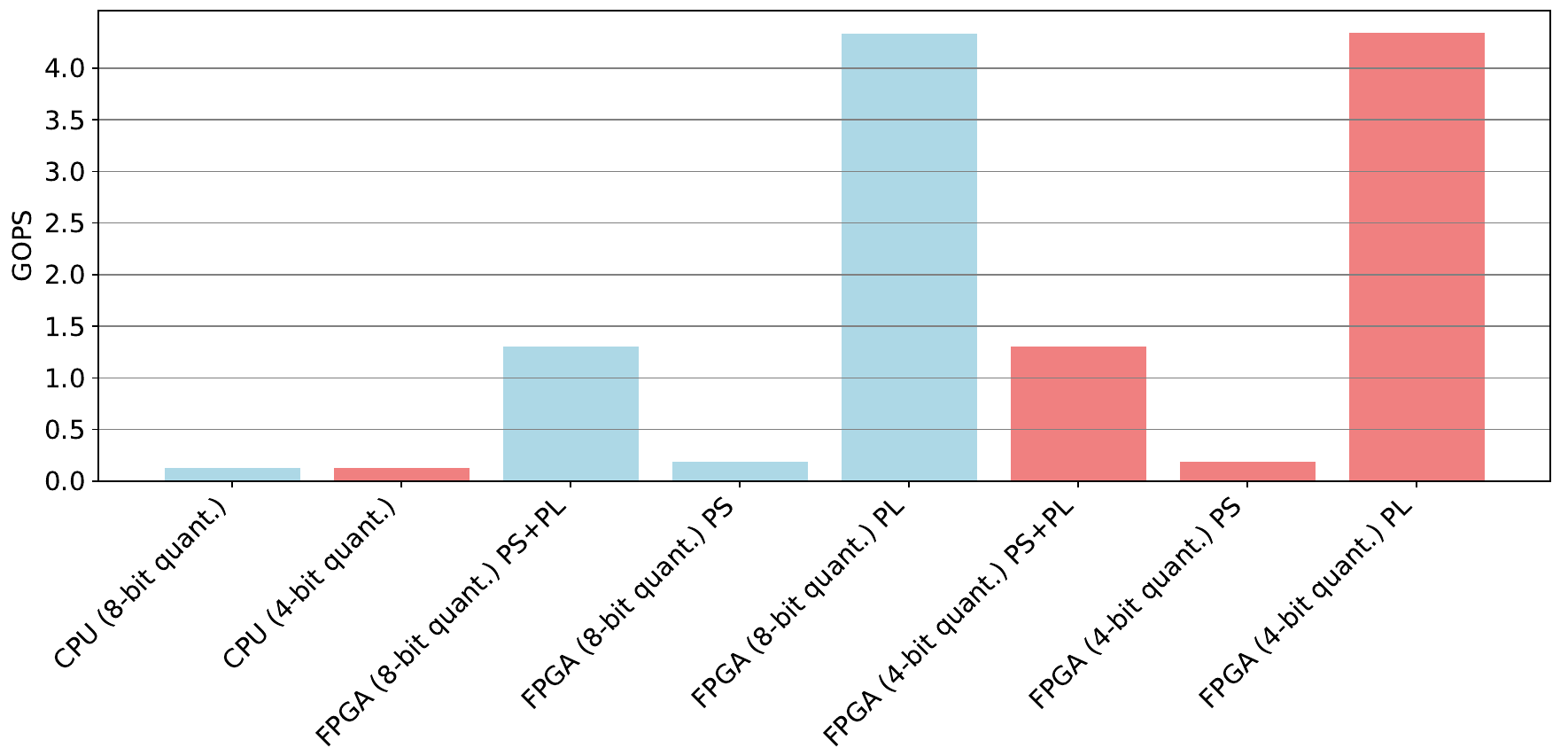}
    \caption{GOPS of CPU and FPGA implementations}
    \label{fig:gops_value}
\end{figure}

Table \ref{tbl:eval_time} also presents GOPS (giga operations per second) values based on the mean execution times.
The results are visualized in Fig. \ref{fig:gops_value}.
The peak performance is 41.2 GOPS when 206 DSP slices are fully-utilized for multiply and add operations at 200MHz, while the measured GOPS value of the PL part is 4.33 GOPS.
The proposed accelerator is implemented as a stream of diverse blocks, such as ODEBlock1, DSBlock1, ODEBlock2, DSBlock2, and MHSABlock, in contrast to classic ResNets that have very uniform structure. 
Considering this diverse structure, our current accelerator is implemented as a modular manner.
In this case, it would be possible to bring the measured GOPS value closer to the peak GOPS value by reducing the number of DSP slices by reusing the same DSP slices for all the blocks.
Also, it would be possible to further reduce the number of
DSP slices for the 4-bit implementation by adopting an aggressive DSP
packing technique.
Such architectural exploration and optimization are our future work.

\subsubsection{Energy efficiency}
The power consumption of the proposed FPGA implementation is evaluated based on power reports from Xilinx Vivado.
For both the 8-bit and 4-bit quantized models, the power consumption in the PS part is 2.64W, while that of the proposed IP core is 1.02W.
Considering that the inference times of the 8-bit and 4-bit quantized models are accelerated by 9.85$\times$,
the energy efficiency is improved by 7.10$\times$.

\subsubsection{Performance comparison against existing work}\label{sssec:perf_comp}

In \cite{Okubo23}, a hybrid model that combines MHSA and Neural ODE based CNN is implemented on ZCU104, though only the MHSA part is accelerated by the PL part.
The performance gain of the MHSA part is only 2.63$\times$ compared to the software counterpart \cite{Okubo23}.
In \cite{Kawakami22}, a Neural ODE based CNN is implemented on the same platform.
The performance gain without the pre- and post-processing is 27.9$\times$ \cite{Kawakami22} and its fps (frames per second) is 16.6.
In this work, on the other hand, the performance gain of the MHSA and CNN parts without the pre- and post-processing is 34.01$\times$ and its fps is 23.0, resulting in a higher performance compared to these works on the same FPGA platform.
Although our targets are resource-limited FPGA devices, ViT variants are accelerated by Xilinx Alveo U200 platform \cite{Marino24} and their ME-ViT achieves 11.99 to 23.98 fps for larger images.

\subsubsection{Practical implications}

Reducing the parameter size while preserving the accuracy
is challenging while it is usually required in resource-limited edge
computing.
Below are our practical implications toward this challenge which were
obtained through the experiments:
\begin{enumerate}
\item Our ablation studies show that the DS+Block3 model where the
  8-bit quantization is applied to DSBlocks and MHSABlock strikes a
  good balance between the accuracy and parameter size.
\item Applying DSC to DSBlocks can further reduce the parameter size
  with small accuracy loss, showing a favorable trade-off between them.
\item Our quantized models are first initialized with parameters of
  the baseline floating-point model and then fine-tuned with a smaller
  learning rate.
  The LLT-based quantization method can be implemented on the FPGA with
  small overheads while mitigating the accuracy loss.
\end{enumerate}

\section{Conclusions}\label{sec:conc}
\subsection{Summary}

Transformers, featuring the highly expressive MHSA mechanism, have demonstrated remarkable accuracy in the field of image processing among recent AI models.
This paper focuses on a hybrid model that combines Transformer and CNN architectures.
Below are our key contributions:
\begin{enumerate}
\item Considering the resource constraints of edge devices, we employed Neural ODE, an approximation technique inspired by ResNet, to design a tiny CNN-Transformer hybrid model with 95.1\% less parameters than ResNet50. The parameter size of the proposed model is the smallest among the state-of-the-art lightweight models while it offers a comparable accuracy.
\item We proposed the resource-efficient FPGA design of the tiny Transformer model tailored for edge devices. We employed a recently-proposed LUT-based quantization technique and conducted the ablation studies to investigate the impact of quantization on different parts of the proposed model.
\item The entire feature extraction part is implemented on a modest-sized FPGA to accelerate inference speed. As a result, the FPGA implementation achieved a 34.01$\times$ faster runtime without pre- and post-processing and a 9.85$\times$ faster runtime for the overall inference compared to an embedded CPU, resulting in a 7.10$\times$ higher energy efficiency.
\end{enumerate}
A demonstration video of the proposed FPGA implementation on ZCU104 board is available on YouTube \footnote{https://youtu.be/jldJGnSYfVQ?si=Ss0PWq2i17hB2Q6S}.

\subsection{Limitations and future direction}

The objective of this work is to demonstrate our proposed lightweight hybrid model combining Neural ODE and MHSA on resource-limited FPGA devices for edge computing.
Thanks to the adoption of Neural ODE and quantization, the parameter size is significantly reduced and thus the proposed model fits within the limited on-chip memory, eliminating unnecessary data transfer overhead.
On the other hand, as mentioned in Sec. \ref{sssec:perf_comp}, the
performance of the proposed IP core is not always higher than those for
larger FPGA devices \cite{Marino24}.
We believe that using larger FPGAs would allow more design choices and optimizations, potentially leading to greater improvements on the speed and accuracy.
While basic forms of the attention mechanism and Neural ODE are employed in this paper, it is possible to further refine the proposed model in terms of computational efficiency and accuracy, as discussed in Sec. \ref{sec:related}.
In addition, in this paper the proposed model was implemented as a sequence of multiple blocks in the PL part.
As an alternative approach, it can be implemented as a single configurable block.
Although the single block implementation is more complex and challenging, it would be beneficial from a resource efficiency viewpoint, enabling a performance improvement in cooperation with more aggressive parallelization.


\end{document}